\newcommand{\Condicionada}{\Sigma \left[ X|X_{meas} \right]}
\newcommand{\CorrelacionCruzada}{\Sigma \left[ X,X_{meas} \right]}
\newcommand{\CorrelacionCruzadaTras}{\Sigma \left[X_{meas},X \right]}
\newcommand{\AutocorrelacionBase}{\Sigma \left[ X,X \right]}
\newcommand{\AutocorrelacionMuestra}{\Sigma \left[ X_{meas},X_{meas}  \right]}
\newcommand{\Entropia}{H\left[ X|X_{meas}\right]}

\documentclass[final,5p,times,twocolumn]{elsarticle}

\usepackage{amssymb}
\usepackage{amsmath,mathtools}
\usepackage{hyperref}
\usepackage{graphicx}
\usepackage[table,xcdraw]{xcolor}
\usepackage{booktabs}
\usepackage{algorithm}
\usepackage{algpseudocode}

\DeclareMathOperator*{\argmax}{argmax}

\newcommand\red[1]{\textcolor{red}{#1}}
\newcommand\changes[1]{\textcolor{black}{#1}}

\usepackage{todonotes}
\journal{Applied Soft Computing}

\begin{document}

\begin{frontmatter}

\title{Censored Deep Reinforcement Patrolling with Information Criterion for Monitoring Large Water Resources using Autonomous Surface Vehicles}
%\title{Censored Deep Reinforcement Patrolling in Water Resources using Autonomous Surface Vehicles: an Entropy-Information Perspective}

\author[inst1]{Samuel Yanes Luis}

\affiliation[inst1]{organization={Department of Electronics, University of Seville},%Department and Organization
            addressline={Av. de Los Descubrimientos s/n}, 
            city={Seville},
            postcode={41003}, 
            state={Seville},
            country={Spain}}

\author[inst1]{Daniel Gutiérrez-Reina}

\author[inst1]{Sergio Toral Marín}

\begin{abstract}
%% Text of abstract
\changes{
Monitoring and patrolling large water resources is a major challenge for conservation. The problem of acquiring data of an underlying environment that usually changes within time involves a proper formulation of the information. The use of Autonomous Surface Vehicles equipped with water quality sensor modules can serve as an early-warning system agents for contamination peak-detection, algae blooms monitoring, or oil-spill scenarios. In addition to information gathering, the vehicle must plan routes that are free of obstacles on non-convex maps. This work proposes a framework to obtain a collision-free policy that addresses the patrolling task for static and dynamic scenarios. Using information gain as a measure of the uncertainty reduction over data, it is proposed a Deep Q-Learning algorithm improved by a Q-Censoring mechanism for model-based obstacle avoidance. The obtained results demonstrate the usefulness of the proposed algorithm for water resource monitoring for static and dynamic scenarios. Simulations showed the use of noise-networks are a good choice for enhanced exploration, with 3 times less redundancy in the paths. Previous coverage strategies are also outperformed both in the accuracy of the obtained contamination model by a 13\% on average and by a 37\% in the detection of dangerous contamination peaks. Finally, these results indicate the appropriateness of the proposed framework for monitoring scenarios with autonomous vehicles.
}

\end{abstract}

%%Graphical abstract
%\begin{graphicalabstract}
%\includegraphics{Figures/GraphicalAbstract.pdf}
%\end{graphicalabstract}

%%Research highlights
%\begin{highlights}
%\item \textbf{Research highlight 1:} A Deep Reinforcement Learning approach to solve the Informative Path Planning and Informative Patrolling problem using Autonomous Surface Vehicles in big water resources.
%\item \textbf{Research highlight 2:} A Censored Q-Function to deal with collisions for the Informative Path Planning and Informative Patrolling task.
%\item \textbf{Research highlight 3:} A formulation of a temporal-dependent informative patrolling, the design of a tailored state representation and reward for the policy optimization.
%\item\textbf{Research highlight 4:} A comparison of performance between the proposed approach and other algorithms and heuristics to validate its utility in the real problem.
%\end{highlights}

\begin{keyword}
Deep Reinforcement Learning \sep Autonomous Surface Vehicles \sep Information Gathering \sep Environmental Monitoring \sep Patrolling
\PACS 0000 \sep 1111
\MSC 0000 \sep 1111
\end{keyword}

\end{frontmatter}

%% \linenumbers

%% main text
\section{Introduction}

\label{sec:Introduction}

% 1.A The contamination problem

Water resources are fundamental to both human life and the economic development of communities. Large water bodies, such as rivers, reservoirs, and also lakes, are not only important for direct consumption but are the core of agriculture. Therefore, the conservation of these is vital for subsistence in many parts of the world. However, despite their importance, 80\% of wastewater from cities and industries is discharged untreated into sea and rivers \footnote{\url{https://www.un.org/en/global-issues/water}}. This situation has dramatic consequences for ecosystems. 

\begin{figure}[ht!]
\centering
\includegraphics[width=0.8\linewidth]{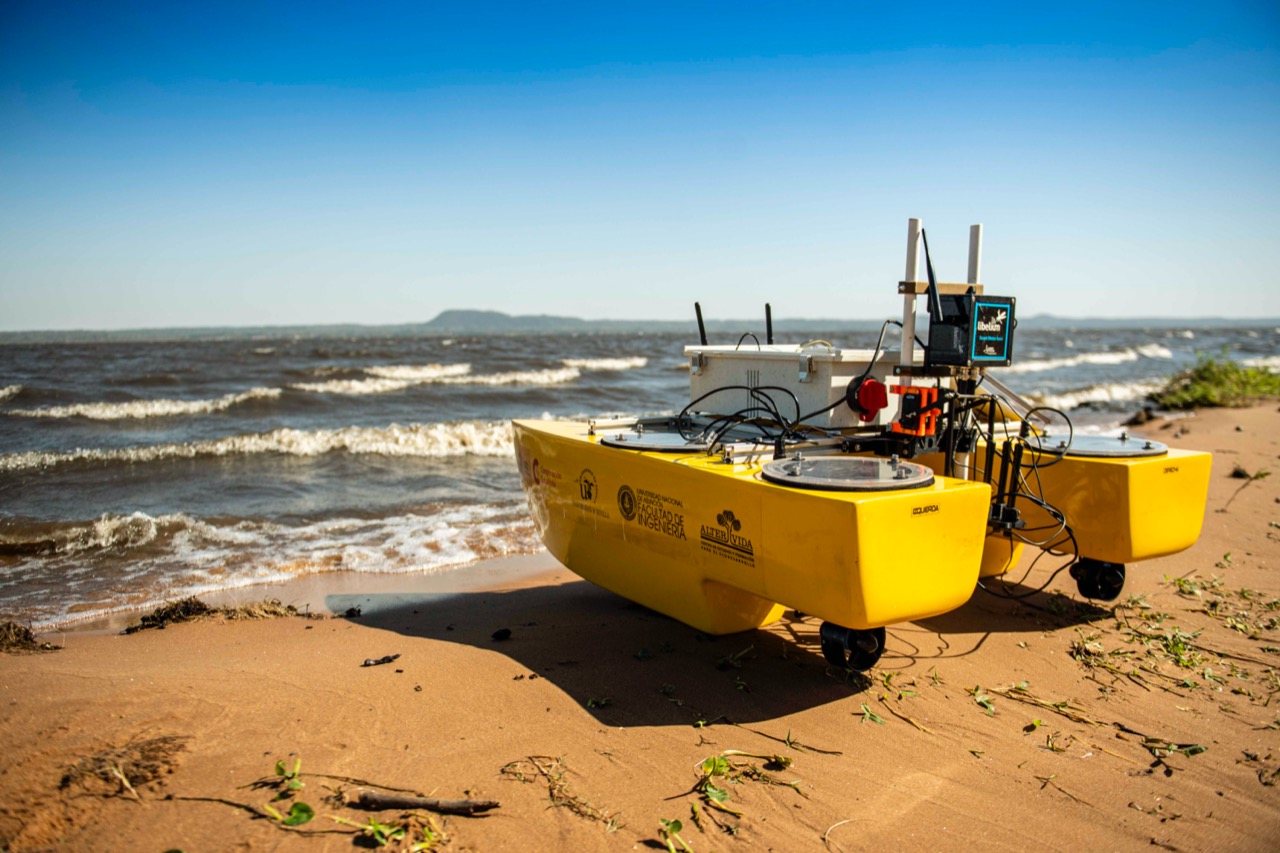}
\caption{ASV prototype for the monitorization of lake Ypacaraí (Paraguay). The ASV is equipped with water quality sensor modules, GPS and communications for remote planning.}
\label{fig:prototype}
\end{figure}

\changes{A very frequent problem in situations of uncontrolled discharges of nutrients is the accelerated eutrophication of waters and the appearance of blue-green bacteria colonies. Whenever the water is nourished with ammonium and nitrites from agricultural fertilizers or human and animal fecal remains, large colonies of cyanobacteria emerge and populate the surface with a greenish mantle. The thick blooms prevent the sun from hitting the underwater flora, which is responsible for oxygenating the waters. As a result, the fish die of anoxia. Thus, the waters become muddy, with pestilential odors and a very poor quality for bathing, fishing, or human consumption. This is the situation in Lake Ypacaraí, Paraguay's largest water resource. The lake, once a tourist enclave, is now populated by colonies of blue-green algae with all its consequences. The situation is similar in places like Mar Menor (Spain) \cite{MarMenorArticle} or Lake Erie (USA) \cite{LagoEireArticle}. All these places have in common that they are very large extensions of water, and that it is difficult to obtain their updated contamination status, in terms of the relevant physico-chemical and water quality (WQ) variables (pH, dissolved oxygen, nitrites or ammonia concentration, etc.). Any solution to this contamination problem requires a comprehensive monitoring process. Monitoring is defined here as the sufficient and homogeneous collection of biological WQ data with which to form a physico-chemical model to serve as an early warning system for pollution peaks. However, this task is difficult due to the large extension size of the ecosystems under monitoring. Consequently, manual missions to measure the WQ are tedious and costly.} %It is necessary to embark a human team, which increases the operational costs, the duration of the missions, and is usually risky in toxic environments. 
Moreover, installing a fixed sensor grid is a suboptimal solution as sampling locations cannot be varied and multiple battery replacements could be even more expensive than human-conducted missions. The use of Autonomous Surface Vehicles (ASVs) equipped with WQ sensor modules has been gaining momentum lately for some similar applications (see Fig. \ref{fig:prototype}) \cite{Sanchez-Garcia2018}. These low-cost electric vehicles allow continuous sampling with routes that can be adapted according to various optimality criteria \cite{arzamendia2019TSP}. %Thanks to advances in robotics and battery autonomy, they are a convenient solution for efficient and adaptive biological monitoring. 
\changes{However, these robotic systems require intelligent perception and decision modules to perform monitoring tasks efficiently. When using these robots, aspects beyond information gathering have to be considered, such as physical battery limitations and nonnavigable terrain constraints. This transforms the monitoring problem into an Informative Path Planning (IPP) problem , which combines the challenge of obtaining the most informative path, and the compliance of ground restrictions and obstacle avoidance. This problem has been previously addressed for an wide set of applications: agricultural characterization \cite{PopovicIPP}, for the generation of water quality models \cite{peralta2021}, the search for gas leaks \cite{Wiedemann}, or the location of contamination sources in radiological environments \cite{radioactivity}. Thus, it is presented as a current problem treated from multiple perspectives within engineering. Our approach can used without loss of generalization to these particular objectives, as the algorithm is model-free and the information formulation can be applied to the reality of these parameters under surveillance apart from the WQ measurement context.}

This paper addresses two cases: the static IPP, in which the variables are assumed to not change throughout a mission, and a temporal case, in which the variables to be measured can change during the mission. The latter case, which can be seen as an Informative Patrolling Path Planning (IPPP), gains importance when we want to measure eventual risks such as spills, new blooms or the time trend of certain pollution variables. This constitutes the base problem of detection and early warning of contamination hazards on long term missions.

To deal with the aforementioned monitoring scenarios, a stochastic framework is defined as it models the information collected at each physical point by the ASV. Assuming that each possible point eligible to be sampled behaves as a Gaussian random variable $p \in X \sim \mathcal{N}(\mu, \sigma)$, a spatial correlation matrix $\Sigma[X, X_{meas}]$ can be defined to model the statistic relationship between samples. This matrix indicates the level of uncertainty of each point, considering the locations where $X_{meas}$ has been sampled. This approach makes use of a Radial Basis Function (RBF) to spatially correlate samples as a function of their adjacency, under the acceptable assumption that physically close samples will be more closely related. Given that, the IPP will consist of sequentially deciding the next physical point at which to take a sample, following the information gain maximization criterion. Following the Information Theory definition of information gain, $\Sigma[X, X_{meas}]$ defines how informative a point is according to the entropy decrease of the model formed by the set of points of the lake $X$ and the sample points $X_{meas}$ \cite{RasmussenBook}. Finally, the IPP problem to be solved here consists of minimizing the total entropy of the scenario $\Entropia$.

The dimension of this sequential problem explodes when the routes are large and movement restrictions are included. In the case of Lake Ypacaraí, large distances must be covered for the routes to decrease optimally the entropy, so the possibilities of movement are almost infinite. In addition, it is necessary for the vehicle to have a reactive policy, capable of adapting to different boundary conditions, i.e., arbitrary route starting points and temporal dependencies. This work proposes the use of Deep Reinforcement Learning (DRL) techniques to find an optimal policy, based on convolutional neural networks (CNNs). This convolutional network is able to interpret a graphical state of the problem and choose the next sample point according to a tailored reward function. This reward function should be designed to evaluate every action in terms of the information gathered. DRL techniques are useful because they do not require a prior model of the system and are able to adapt to a scenario with arbitrary boundary features through model-free interpretation of the state and reward function. These characteristics make DRL a suitable approach to solve the IPP and IPPP. Specifically, Deep Q-Learning (DQL) are applied here because of their ability to estimate the future reward given a state for each possible action in the possible set. To enhance the capability of the classical DQL algorithm, this paper further elaborates on the application of some useful techniques for this IPP problem, that have been proven effective in previous works of the DRL literature \cite{RainbowDQL}, but none in a path planning application: the use of a Prioritized Experienced Replay \cite{schaul2015prioritized}, a Dueling architecture in the \textit{Q} neural network \cite{DuelingQ} and, finally, the use of noisy neural networks \cite{NoisyNetworks} to intrinsically motivate the exploration of the state-action domain. \changes{These publications revealed the high learning capacity of algorithms such as DQL. However, in the application of them to complex problems, there are hard constraints that are known a priori or can be computed in a deterministic way. A difference between the proposed approach and the previous ones is in using the deterministic knowledge of the navigation map to increase the sampling efficiency. This is effectively a slight departure from the initial model-free paradigm of DQL. However, we propose the use of a censoring mechanism that increases the sample efficiency because it provides accurate information on illegal actions that would otherwise have to be inferred by trial and error. This algorithm, the Censoring-DQL, takes the deterministic information of the environment related to the obstacle avoidance task, and provides a way to neglect those actions that incurs in violations of the hard-constraints in navigation. As far as we understand, there is no apparent compromise on learning convergence when this mechanism is used and, moreover, it results in improved training efficiency.}

\changes{The novelty of this work also lies in the formulation of the monitoring problem as a sequential entropy minimization process, using the principles of DRL: The problem is defined as a discrete Markov decision process (MDP) in which the objective is to maximize information acquisition with a finite planning horizon. Thus, it is proposed to use the information gain as the reward that reduces the total uncertainty under the assumption that the water quality parameters behave smoothly and as Gaussian random variables. As explained in \cite{Sutton1998}, it is usually challenging to design the reward function to solve a real problem with as it must reflect the desires of the resulting behavior. In the particular IPP/IPPP case, this tailored reward function serves as a surrogate signal for a general objective: information acquisition, translated into model error reduction, maximal search, etc. This reward-engineering process contributes, to improve data collection systems and to obtain deep policies for decision making in the face of information uncertainty. With an appropriate definition of these learning parameters, the DRL is able to outperform previously proposed algorithms and heuristics, with more efficient and robust informative paths. In previous works, like in \cite{arzamendia2019online}, the patrolling follows a criterion based on the area covered and not on the information, making the paths longer and more redundant. In addition, other similar algorithms, such as \cite{arzamendia2019TSP}, use paths from side to side of the edges, without considering interior paths that do not intersect. Another disadvantage of these heuristics is the inability to act online, since they base the optimization on a static criterion and, therefore, a single solution rather than a policy is obtained.}

In summary, the contributions of this work are: I) A model-free Deep Reinforcement Learning framework to solve the IPP and the IPPP in water monitoring scenarios using an entropy minimization criterion in stationary and nonstationary scenarios. II) The definition of a tailored reward function, a graphical state formulation, and a noisy CNN architecture for policy optimization that is effective in both the IPP and the IPPP scenario.

\changes{This article is organized as follows: in Section 2, there is a brief survey of previous works and other techniques that address similar problems. In Section 3, Materials and Method, the IPP and IPPP problem is formally stated, the algorithm is presented, and the Deep Reinforcement Learning framework described. In Section 4, the results of the training are analyzed and discussed. Finally, Section 5 closes with the conclusions of this work and future lines of work.}

\section{\changes{Previous works}}

The utilization of ASVs is an increasingly common option for autonomous operation in hydrological resources and harbors \cite{Ferreira2009, Yanes2020, peralta2021, peralta2021bayesian, arzamendia2019online, kathen2021informative, luis2021multiagent}. Applications range from bathymetry surveys \cite{Ferreira2009}, pollutant modeling and monitoring \cite{Yanes2020}, or patrolling for early warning . In most applications, Artificial Intelligence or metaheuristic optimization techniques are usually applied to solve complex NP-hard problems: Bayesian optimization \cite{peralta2021bayesian}, genetic algorithms (GA) \cite{arzamendia2019online}, Swarm Intelligence \cite{kathen2021informative}, or DRL \cite{luis2021multiagent}.

For the acquisition of a surrogated model of the environment, in \cite{peralta2021} Bayesian Optimization is used to obtain a model of Lake Ypacaraí itself with a single ASV using Gaussian Processes (GP). In that work, the suitability of different kernels for GP and novel acquisition functions to meet the vehicle battery constraints is deepened. Along the same lines, in \cite{peralta2021bayesian} the algorithm is extended to a multivariate case, where there are different objective functions to fit. Both approaches seek to minimize the regression error by choosing the next point to sample using an acquisition function that balances the exploration/exploitation through uncertainty and GP averaging. However, it does not take into account possible temporal changes of the benchmark functions under monitoring. Moreover, our proposal attempts to find an intensive coverage with a higher number of samples and is not particularly tied to any particular regression method. In \cite{PopovicIPP} the use of aerial vehicles with a camera for environmental characterization is also proposed. This work uses the CEM-ES evolutionary algorithm to minimize entropy in a Kalman Filter model for adaptive modeling. Our work also makes use of the entropic criterion as an indicator of the information obtained, but with a smaller data regime and, again, with an application for temporal patrolling. \changes{A similar criterion is used in \cite{radioactivity}, where the entropy leverages the remained information. Nonetheless, the approach is seen to be problematic when surveying convex scenarios, and its performance is surpassed by other heuristics. In our approach, we present solid results of the utility of the presented algorithm, and, again, an extension to temporal patrolling, which is not addressed in \cite{radioactivity}. Regarding the use of DRL for modeling, in \cite{ViserasDeepIG} it can be found the use of Proximal Policy Optimization with CNNs to minimize the mean square error (MSE) of an objective function with multiple vehicles. This aforementioned work demonstrates that it is possible to find effective multiagent policies with DRL for monitoring. In \cite{Wiedemann}, a similar approach was presented, but for gas leaks localization. This approach used a differential model of the gases to provide a sufficiently accurate model for the deep agent to learn. Our approach, on the contrary, works in a full model-free fashion respect to the information, as it is sometimes hard when not impossible to model the behavior of WQ parameters in such big resources.}

An example of maximizing the area covered in water resource monitoring can be found in \cite{arzamendia2019TSP}. This approach is based on solving a Travel Salesman Problem (TSP) using Eulerian cycles to maximize the unvisited area using a single vehicle with GA. This approach only uses edge-to-edge movements and is limited by the formulation of visitable nodes. Our algorithm proposes to overcome this method by allowing movements in different directions at each instant. Moreover, the information coverage is no longer based on an all-or-nothing basis, but follows the information criterion imposed by the RBF function. Another example of full coverage algorithms is in \cite{KrishnaLakshmanan2020}, where a DRL algorithm for a transformable mobile robot with discrete actions is applied. In this work, the algorithm called Actor-Critic with Experience Replay (ACER) and a visual formulation of the state with CNNs are used in the same way as in our proposal. However, the application again consists of a binary coverage (covered/uncovered regions) with no temporal criterion or useful redundancy factors in the coverage. 

Regarding the particular use of DRL for path planning of vehicles, most of the works focus on simply finding obstacle-free routes \cite{ZHANG2022108194} or low-level control \cite{ZIELINSKI2021107602}. In \cite{ZHANG2022108194}, the Twin Delayed Deep Deterministic policy gradient (T3D) algorithm is implemented for obtaining optimal obstacle avoidance policies for aerial vehicles with distance sensors. \changes{In \cite{ZIELINSKI2021107602}, the DRL is applied in obtaining a low-level tracking policy for an Autonomous Underwater Vehicle (AUV) using visual inputs from a camera. By the means of the TinyYOLO architecture, an convolutional obstacle detection system, the AUV is able to track the desired paths without collisions. Another interesting application of DRL in vehicles can be found in \cite{KOH2020106694}, where a DRL policy is optimized to solve the vehicle routing and navigation task in an interactive and self-adapting manner. The difference in the sense of obstacle-avoidance between such approaches and ours is that the proposed censoring mechanism implies a simplification. The navigation map is known a priori, and the obstacles are observable. Then, the decisions related to the information acquisition are surrogated to the detection mechanism and therefore. With regard to patrolling and surveillance tasks, there are previous coverage applications \cite{Theile2020, Piciarelli2019} that use DRL with a visual formulation of the state. In \cite{Theile2020}, the use of DQL is proposed for the realization of partial coverage routes based on a battery budget with vehicle landing constraints. Our work differs in the sense of the resolution and the state-action domain, which is significantly bigger in our case. The work \cite{Theile2020} is more similar to a low resolution task like \cite{Yanes2020}. In \cite{Piciarelli2019}, an interesting example of nonhomogeneous temporal patrolling can be found, where emphasis is put on going over the most important areas with longer waiting time, in a similar way, but with a discrete redundancy criterion. The proposed work differs from \cite{Piciarelli2019} in the presence of nonnavigable zones. There is also a difference in the acquisition system. Whereas in \cite{Piciarelli2019} a camera is in charge of sensing, the proposed approach uses point samples and generates a model in between.}

In this work, we propose a framework based on DRL and information theory, which, as far as the authors are aware, fills the following research gaps: i) most applications of DRL in vehicles are limited to low-level control or path planning, with results similar to classical algorithms. This work proposes a way to optimize more complex problems, such as entropy minimization for environmental monitoring, with a visual formulation of the process. ii) The proposed framework is able to simultaneously deal with obstacles while minimizing uncertainty, and is useful when information varies temporally or in the presence of outliers.

\section{\changes{Materials and Methods}}

This section introduces the mathematical framework of the monitoring problem and the assumptions taken to model the scenario. Both monitoring cases are considered: a static scenario, where the gathered information does not recover its importance once obtained, and a dynamic scenario, where the unvisited zones increase its uncertainty within time.

\subsection{Entropy framework}

To present the IPP and IPPP problems, we begin with the definition a navigable space $X \in \mathbf{R}^2$, where the vehicle can take WQ samples.  We also define the subset set of sampled locations $X_{meas} \in \mathbf{R}^2$, and a binary matrix $M$ which is a navigability map such that $X = \forall p:=[p_x,p_y] \; |; M(p) = 1$. Now, we define that each visitable point $p$ behaves as a Gaussian random variable of mean $\mu$ and standard deviation $\sigma$ such that $p \approx \mathcal{N}(\mu,\sigma)$. The space $X$ is thus formed as a Multivariate Gaussian Distribution (MGD), where $X \approx \mathcal{N}_n \left( \boldsymbol \mu, \boldsymbol \Sigma \right) $, with $\Sigma$ being the correlation matrix of $X$.

The measure of the correlation between points can be obtained by means of a function that indicates how closely related two variables are in the search space. For this application, a RBF kernel has been chosen, so that the correlation between two variables decreases smoothly and exponentially with the distance between them according to the Eq. \eqref{eq:RBF}. This choice makes sense from the point of view of water resources monitoring since physico-chemical variables usually have a smooth distribution\footnote{\url{https://marmenor.upct.es/maps/}}. Thus, two samples taken in close proximity will presumably be highly correlated and the information they both provide can be redundant. The parameter $l$ will serve to scale how much correlated two measurements $(p,p')$ are with each other and is usually chosen based on prior knowledge of the environment to be monitored \cite{ViserasDeepIG, PopovicIPP} \changes{or intensity of the environment coverage}.

\begin{equation}
RBF(p,p') = exp \left( \frac{(p_x - p'_x)^2 + (p_y - p'_y)^2}{2l^2} \right)
\label{eq:RBF}
\end{equation}

Thus, we can find the correlation matrix of the navigable space conditional on the sampled locations by the element-wise evaluation of Eq. \eqref{eq:RBF}. The conditional correlation matrix $\Condicionada$ will have the expression of the Eq. \eqref{eq:conditioned} \cite{RasmussenBook}.

\begin{equation}
\begin{split}
{\Condicionada} = & \\{\AutocorrelacionBase} - {\CorrelacionCruzada} \times & {\AutocorrelacionMuestra}^{-1} \times {\CorrelacionCruzadaTras}^{-1}
\end{split}
\label{eq:conditioned}
\end{equation}

Now, the monitoring objective involves decreasing the entropy associated with the conditional correlation. The information entropy $\Entropia$ gives a measure of the uncertainty about the monitoring domain and the randomness of a sample at an arbitrary point in that space. The lower the entropy, the more confidence one has about the scenario. In this sense, under the assumption that the monitoring space $X$ is an MGD, the entropy gives a lower bound on the error of an arbitrary estimator $\hat \mu$ \cite{InformationTheory}, as given in the Eq. \eqref{eq:lowerbound}.

\begin{equation}
\mathrm{E}\left[ (\mu - \widehat{\mu})^{2} \right] \geq \frac{1}{2 \pi e} e^{2 H(\mu | X_{meas})}
\label{eq:lowerbound}
\end{equation}

Finally, the entropy can be calculated as \cite{RasmussenBook}:

\begin{equation}
\Entropia = \frac{1}{2} \log(|\Condicionada|) + \frac{dim(X)}{2} \log(2 \pi e)
\label{eq:entropy}
\end{equation}

\begin{figure}[t]
\centering
\includegraphics[width=0.9\linewidth]{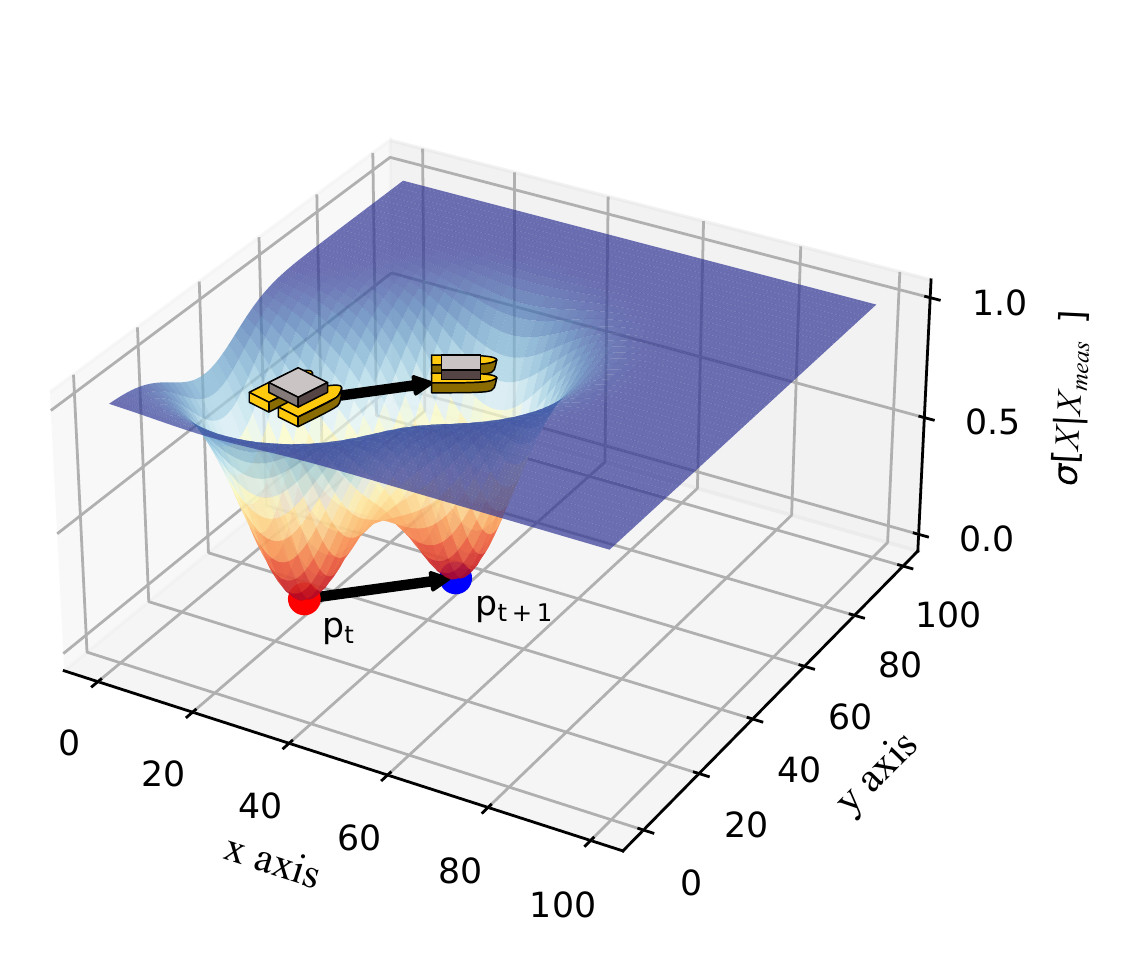}
\caption{Uncertainty conditioning process. When we go from $p_t$ to $p_{t+1}$ and incorporate a new sample to $X_{meas}$ with the Eq. \eqref{eq:conditioned}, we can obtain a new $\Condicionada$ matrix. It is observed that the standard deviation associated with $\Condicionada$ becomes zero at the sample points (assuming no sampling noise) and in their environments according to the RBF function. The entropy $\Entropia$ of the process decreases accordingly.}
\label{fig:conditioning3d}
\end{figure}

An example of how the min-max normalized standard deviation values $\sigma$ of $\Condicionada$ (its diagonal) decrease when starting from a sample at $p_t$ to $p_{t+1}$ can be seen in Figure \ref{fig:conditioning3d}. In the vicinity of those sample points, $\sigma$ is not 1.0 at all since a smooth correlation is assumed with the RBF function.

\subsection{Temporal-dependent entropy}

In a more general case, where the variables to be monitored change during a mission, it is necessary to readjust the entropy formulation to include such factor. To do this, a monotonically increasing time-dependent function $f(t)$ is included in the correlation matrix $\Condicionada$ calculus, such that the uncertainty of the measurements increases as a function of the time that has passed since each location $X_{meas}$ was visited. Thus, the definition of $X_{meas}$ is expanded to include the instant of visit of each sample $t_{meas}$. Similar to \cite{DynGPBranke}, the correlation matrix of samples $\AutocorrelacionMuestra$ is calculated by adding the temporal term to the diagonal:

\begin{equation}
\AutocorrelacionMuestra = RBF(X_{meas},X_{meas}) + \tau \left( t - t_{meas}\right)^2 \mathbf{I}
\label{eq:temporal_entropy}
\end{equation}

This is analogous to increasing the noise of each particular sample quadratically with time. Old samples do not contribute to decreasing entropy, and over time, unvisited areas recover their original entropy. In the Eq. \eqref{eq:temporal_entropy}, the value $\tau$ is a forgetting factor that modulates how fast the correlation in each zone regenerates. Thus, with this formulation, decreasing entropy involves not only visiting areas never visited before, but cyclically patrolling the areas to prevent entropy from increasing. 

\subsection{Deep Q-Learning}

In this section it is presented the foundations of the selected DRL algorithm, Double Deep Q-Learning \cite{Mnih2015}. The different modules of the reinforcement learning paradigm are presented, such as the state representation, reward function, Deep \textit{Q}-function, and the variations from the original approach in \cite{Mnih2015} that enhance the learning in this scenario.

Deep reinforcement learning (DRL) is a methodology for solving sequential problems such as the proposed one \cite{Yanes2020, luis2021multiagent}. In DRL, a Markov Decision Problem (MDP) is described so that an agent (vehicle), which is in a state $s_t$, takes an action $a_t$ according to a policy $a_t = \pi(s_t)$. The environment processes the action and returns the next state $s_{t+1}$ and a reward value $r$. To solve a MDP one has to find the optimal policy $\pi^*(s)$ that maximizes the discounted reward in a control horizon $T$:

\begin{equation}
\pi^*(s_t) = \max_{\pi(s_t)} \sum_{t=0}^{T} \left[ R(s_t,a_t = \pi(s_t)) \right]
\end{equation}

Deep Q-Learning (DQL) algorithms within DRL are based on finding an action-value function $Q(s,a; \theta)$, represented by a Deep Neural Network (DNN) with $\theta$ being the trainable parameter, which estimates the discounted future reward, given a state $s$ and for each possible value of actions $a_t \in A$ according to the Eq. \eqref{eq:Qfunction}. The value of $\gamma \in [0,1)$ discounts the value of the reward function $r(s_t,a_t)$ over a finite decision horizon.

\begin{equation}
Q(s_t,a_t; \theta) = \mathbb{E} \left[ r(s_t,a_t) + \gamma \max_{a'} Q(s_{t+1},a'; \theta) \right]
\label{eq:Qfunction}
\end{equation}

The \textit{Q} function is estimated by collecting experiences $E \sim \; <s_t,a_t,s_{t+1},r_t>$, storing them in a experience-replay buffer, and training the DNN in batches of size $|\mathcal{B}|$. The training procedure is based on updating the network parameters $\theta$ in the direction of the descending gradient of the loss function. This loss function is computed as the Time Difference ($TD$) error between the chosen action of $E$ and the best predicted values of a discounted target function $Q^*(s',a,\theta')$ (see Eq.  \eqref{eq:loss}). Thus, the parameters $\theta$ of the network are updated by taking an stochastic gradient descent step with a learning rate of $l_R$.

\begin{equation}
\mathcal{L} =  \left[ \underbrace{Q(s_t,a_t; \theta) - \left( r_t + \gamma \max_{a'} Q^*(s_{t+1},a'; \theta')\right)}_{TD} \right]^2
\label{eq:loss}
\end{equation}

\noindent Through trial and error, the \textit{Q} function is adjusted to the real value and the agent learns to select the best actions. The target function is periodically updated with the value of the \textit{Q} function itself \cite{Mnih2015} to improve the stability of the training. To have a bias-free estimate of the \textit{Q} function, it is necessary to collect both good and bad experiences. The balancing of the exploration of the action-state domain and the exploitation of what is learned is usually left to the DQL by a $\epsilon$-greedy policy. This policy balances exploration and exploitation by taking a random action to explore with $\epsilon$-probability. The value of $\epsilon$ starts high at the beginning of training and decays as learning is completed to take advantage of the knowledge gained in the early stages.

\subsection{\changes{Proposed framework}}

The proposed problem is based on sequentially choosing $p_{t+1}$ that provides the most information from the perspective of entropy reduction. The formulation of the actions, the state, the deep policy architecture, and the reward function are described below. For a complete understanding of the system, see Fig. 1, which visually describes the framework based on the DRL Censoring method, its inputs and outputs.

\begin{figure*}[t]
\centering
\includegraphics[width=0.8\linewidth]{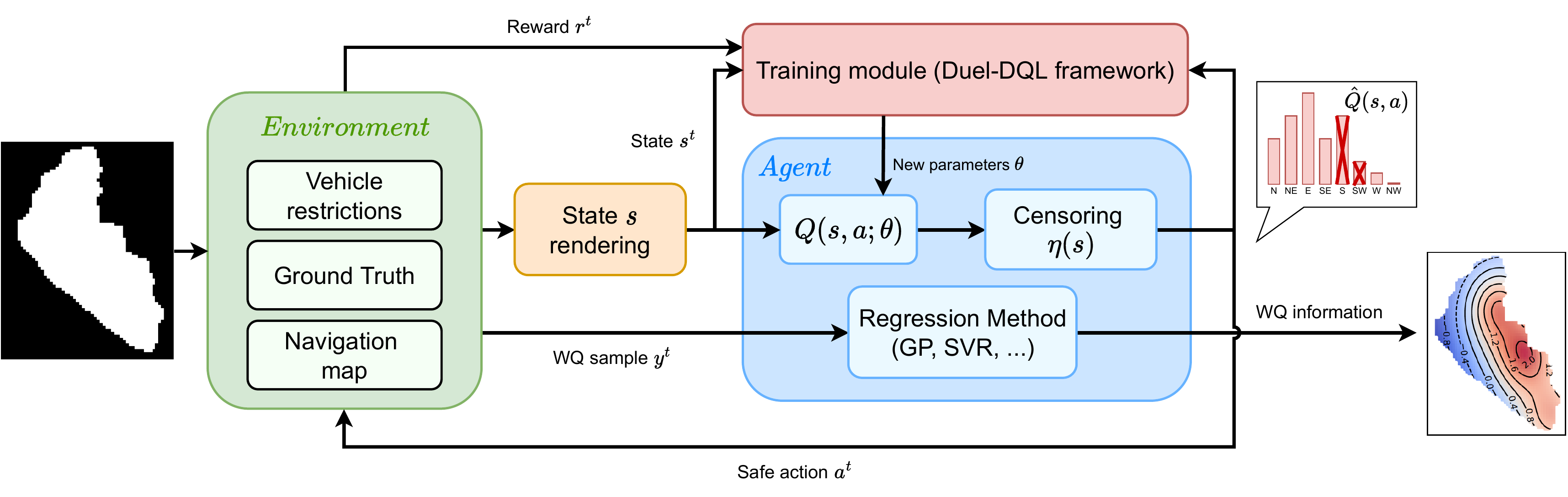}
\caption{This diagram describes the general operation of the system. The algorithm is given a simulation environment based on plausible WQ data and a navigation map. The agent uses a deep policy $Q(s,a)$ to decide the direction of movement in which to take the next sample. The censoring mechanism prevents invalid (hard-constrained) actions. Simultaneously, a WQ model is being shaped by a regression method (Gaussian Processes, Support Vector Regressor, etc).}
\label{fig:diagram_of_framework}
\end{figure*}

\subsubsection{Actions}

\begin{figure}[t]
\centering
\includegraphics[width=0.8\linewidth]{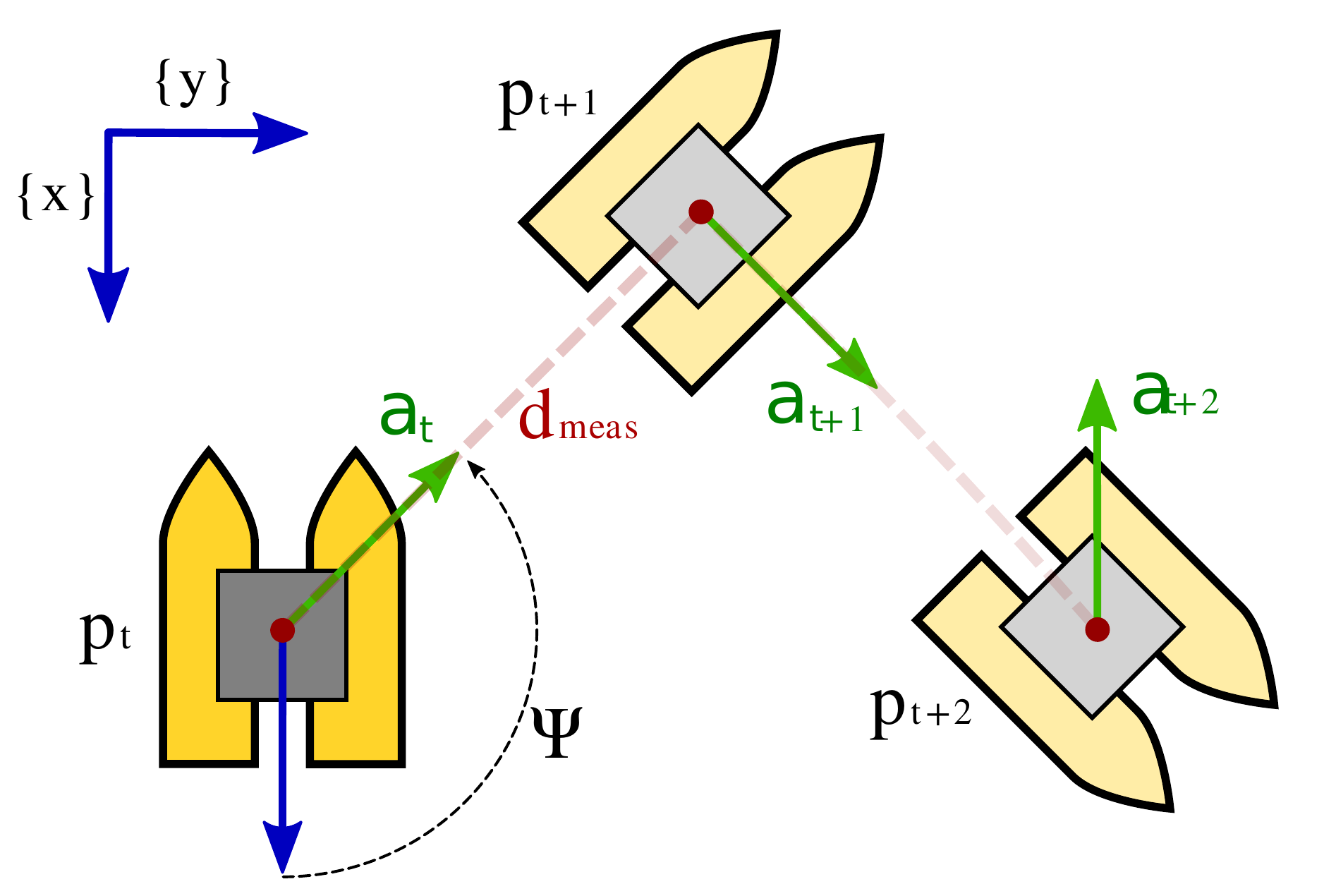}
\caption{\changes{The ASV is moves in 8 possible directions $\Psi$. Those directions are fixed respect to an inertial reference frame parallel to the navigation map $M$. Every action implies a travel in the selected direction for a distance $d_{meas}$. Therefore, every point $p_t$ is separated from the previous one $p_{t-1}$ by a distance of $d_{meas}$.}}
\label{fig:movement}
\end{figure}

\changes{A discrete formulation of the action space is chosen to reduce the number of possible action-state combinations. Discretizing the action space in this way has proven to be sufficient in previous works \cite{PopovicIPP} and allows for better convergence of policies \cite{ComparisonYanes}. Thus, the ASV agent can choose from up to 8 possible actions that will result in a movement in 8 different directions with respect to a fixed reference system parallel to the axes of the navigation map $M$. The possible angles of direction $Psi$ follow the bearings of a compass $A:=$[S, SE, E, NE, N, NW, W, SW]. Note that the order of these angles does not intervene at all in the algorithm. An intuitive order has been chosen but, for the sake of the algorithm, an arbitrary order could have been chosen. Therefore, each action $a$ leads the ASV to a movement in that direction until it reaches a point at a distance of $d_{meas}$ with respect to the previous point. In Fig. \ref{fig:movement} it is depicted the process of movement within its reference frame. Regarding the dangerous movements, any action that generates a waypoint outside the navigable areas is neglected and will not cause any movement. For any effective action, a sample of the WQ values is taken and the new position is updated in the set of visited zones such that $X_{meas}^{t+1} \leftarrow X_{meas}^{t} \cup p_{t+1}$. With this, the covariance matrix is recomputed according to Eq. \eqref{eq:conditioned} and the next action can be queried to the policy $\pi(s)$. The number of possible actions is determined by the length of the path $D$, which will depend on the problem we are solving and the ASV battery budget.}

\subsubsection{State}

\begin{figure}[t]
\centering
\includegraphics[width=0.9\linewidth]{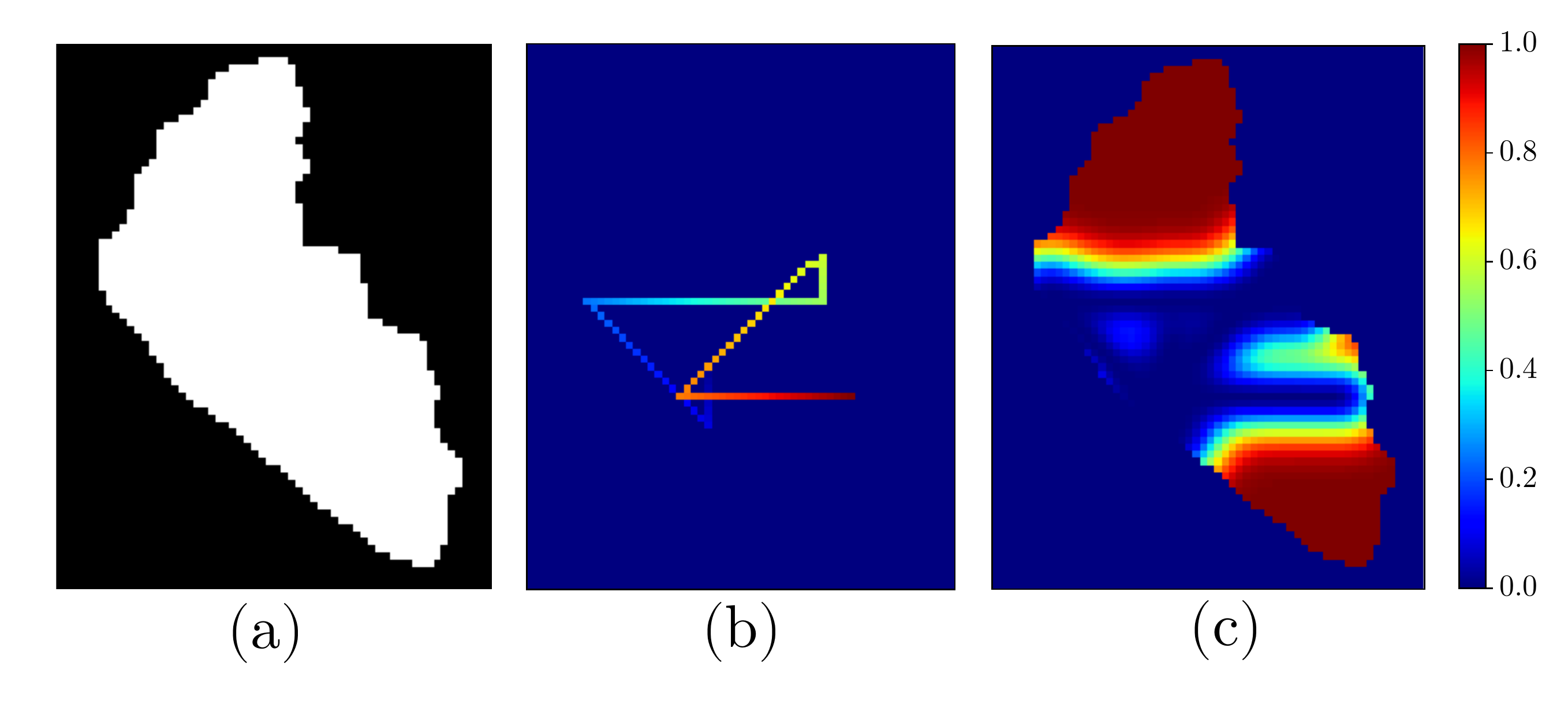}
\caption{The 3 channels of the state: (a) corresponds to the navigation map of the Ypacaraí Lake in Paraguay; (b) corresponds to the path traveled so far; (c) corresponds to the uncertainty map given by $diag(\Condicionada)$. }
\label{fig:states}
\end{figure}

The state $s_t$ represents the observable information of the scenario at a certain instant $t$. To fulfill the Markovian hypothesis, the future reward should depend exclusively on the current state and the action taken. Consequently, the state of the problem must be chosen carefully, including all necessary a priori information for entropy minimization. \changes{As the IPP is a problem with physical constraints dependent on the current position of the navigability map, in addition to the information collected, an image-like state with three channels is proposed: i) the binary navigability map, ii) a map of the path followed so far, and iii) an uncertainty map in which each cell has a value equal to the value of $\sigma$ according to the conditional correlation matrix $\Condicionada$ (see Figure \ref{fig:states}). This last state will depend on the kernel hyperparameters as it is mentioned in Section 3.1. and will change according to Eq. \eqref{eq:conditioned}. This way, the state $s = [s_1, s_2, s_3]$ is discretized into three images of $75 \times 60$ pixels. These images represent all the information available for the scenario, which makes this problem a Fully Observable MDP (FOMDP)} To avoid large input values to the Deep \textit{Q}-network, the tree images are minmax-normalized. This way, every pixel value of the state is within $[0,1]$.

It is worth mentioning that, to avoid violating the Markovian assumption, the path traveled by the ASV up to the current instant is included as part of the image (see Figure \ref{fig:states}b). In order to distinguish the old positions from the current one, the oldest of the positions will have a value of 0 and the current one will have a value of 1. 

\begin{figure*}[t]
\centering
\includegraphics[width=0.9\linewidth]{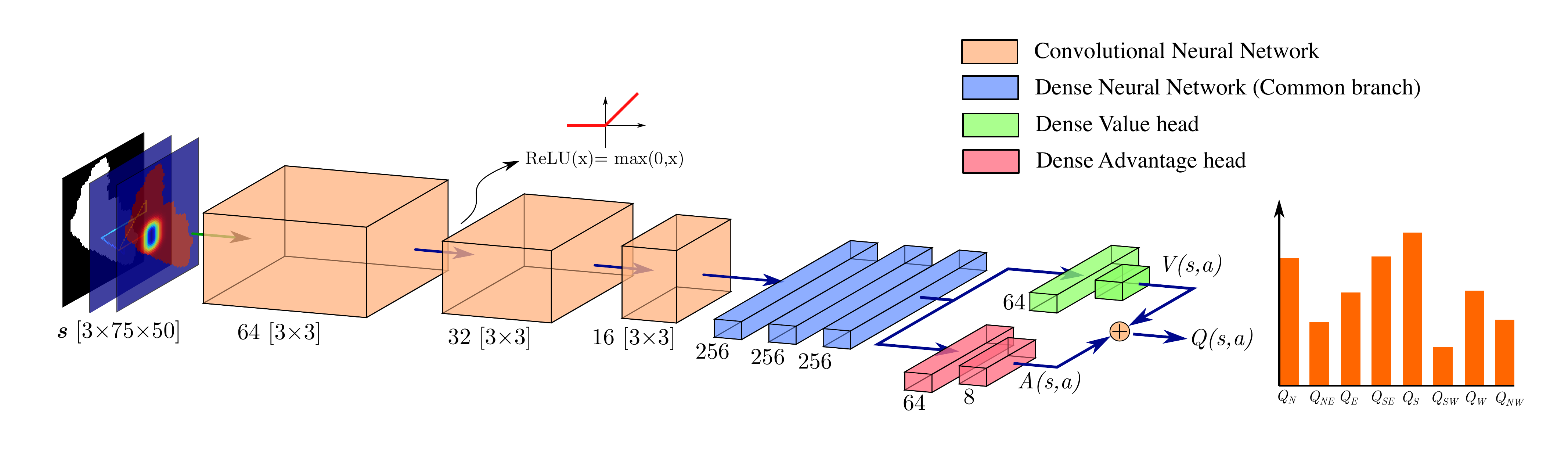}
\caption{\changes{CNN that represents the $Q(s,a;\theta)$ function. The dueling-network processes the state to obtain the advantage value $A(s,a)$ and the stave-value $V(s)$. All the activation layers corresponds to ReLU  function. The output of the network will be 8 \textit{Q}-values corresponding to expected discounted accumulated reward for taking every action $a \in A$ in the current state $s$ (the input of the network). Note that these values does not take in account any invalid (hard-constrained) actions.}}
\label{fig:network}
\end{figure*}

\subsubsection{Noisy Neural Architecture}

To represent the function $Q(s,a)$, we propose the use of a CNN to process the state as an image. A Dueling architecture has been designed. With this architecture, we seek to compute two reward estimators: on the one hand, the value function $V(s)$, which represents the estimated future reward in the state $s$. On the other hand, the advantage function $A(s,a)$, which represents the estimated future reward for each action $a$ with respect to the value of $V(s)$:

$$A(s, a) = Q(s, a) - V(s)$$

Thus, the calculation of $Q(s,a)$ is performed such that:

\begin{equation}
Q(s,a) = V(s,a) + A(s,a) - \frac{1}{|A|} \sum A(s,a)
\label{eq:advantage}
\end{equation}

\changes{This technique allows to better generalize the learning of actions, in the presence of similar $Q(s,a)$ states \cite{DuelingQ}. The proposed network is composed by two parts: the first is a visual feature extractor in charge of transforming the spatial relationships of the state (images) into a feature vector using CNNs. The feature extractor is composed of 3 consecutive convolutional blocks, with 64, 32 and 16 filters each, of size 3x3. Then, 3 layers of fully connected neurons are used. Finally, the 256 output values are processed in parallel by a value head $V(s)$ and an advantage head $A(s)$ with a size of 64 neurons each. Using Eq. \eqref{eq:advantage}, the final values of $Q(s,a)$ are obtained. Every activation layer corresponds to a Rectified Linear Unit (ReLU), except for the output layer, which does not have any (see Figure \ref{fig:network}) to effectively represent the real values of \textit{Q}.}

\begin{figure}[t]
\centering
\includegraphics[width=0.6\linewidth]{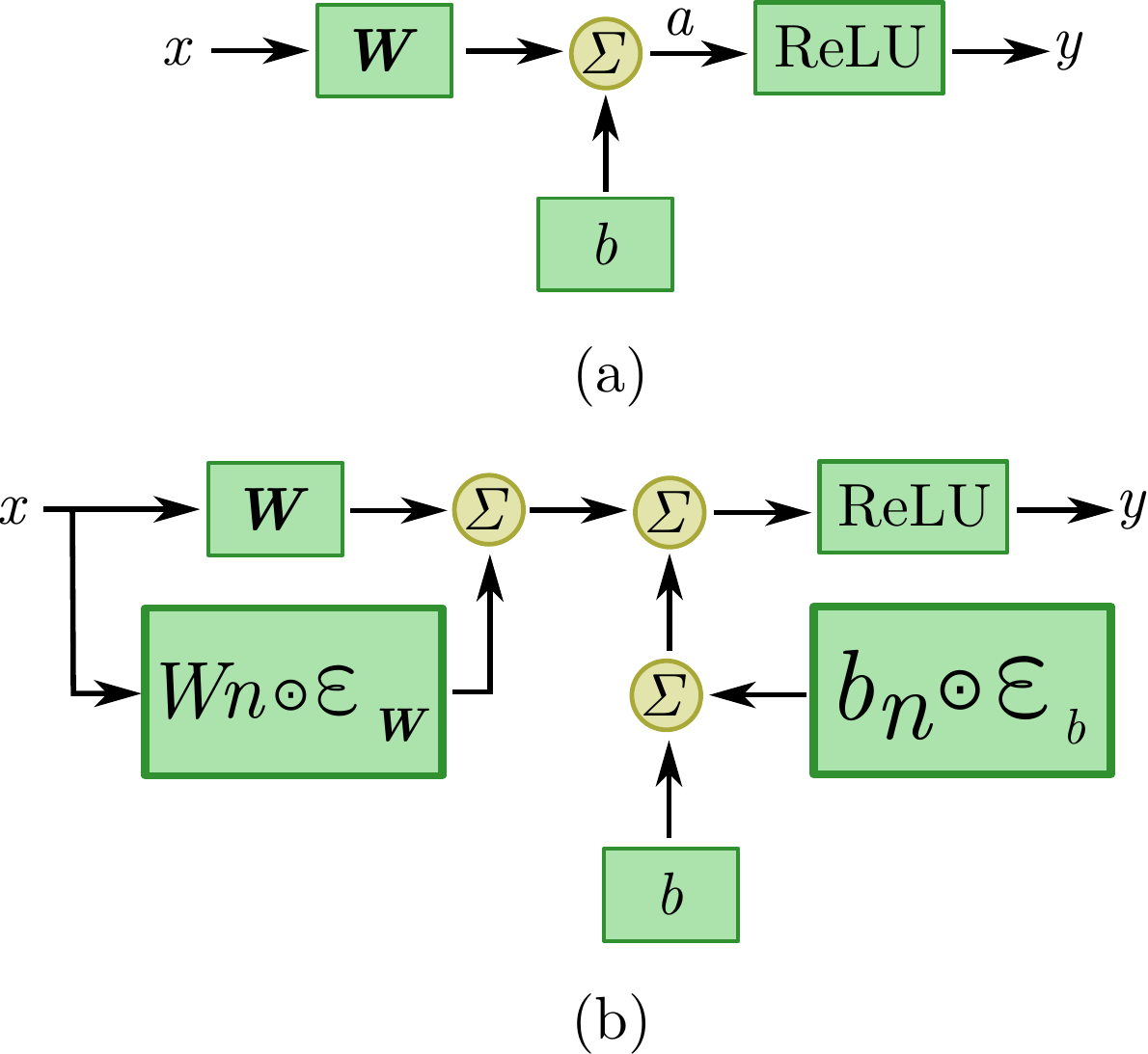}
\caption{In (a), it is depicted a classic neuron with ReLU activation function. In (b), a noisy neuron is presented. It is added to the weights and biases a stochastic value weighted by two trainable parameters $(W_n,b_n)$.}
\label{fig:noisy}
\end{figure}

Furthermore, it has been also implemented a technique useful in the efficient exploration of the state-action domain called Noisy Network \cite{NoisyNetworks}. The classic neurons are replaced by noisy neurons that vary their value at each evaluation in a stochastic manner. To the weights of each neuron $(W,b)$ are added the parameters $(W_n, b_n)$ which are trainable and are weighted with random values $(\epsilon_W, \epsilon_b)$ taken from Gaussian distributions ($\xi_W, \xi_b$) (see Figure \ref{fig:noisy}). In this way, the neural network introduces an intrinsic method of exploration. This is different from classic dropout layers , implemented to avoid over-fitting. The noisy strategy embeds the exploration into the policy and lets the deep agent decide the level of noise in its actions as it learns. This method can be interpreted as a form of evolutionary strategy in which the exploration policy itself is embedded within the agent's action policy. Similar studies, such as \cite{ZHANGDiffEvol}, have shown that these techniques are able to return better solutions thanks to their enhanced exploration capability. 

 \subsubsection{Prioritized Experience Replay}
 
 A fundamental part of DQL is the so-called Experience Replay (ER). As it is mentioned before, the ER consists of sampling batches of previous experiences, saved in a memory buffer $\mathcal{M}$ as they occur, to fit the $Q(s,a)$ function. It is commonplace to use uniform random sampling to batch the experiences to avoid correlated values that will lead the NN to overfitting easily. Nonetheless, with this method, every previous experience has the same probability to be sampled, without considering any further information like the knowledge $Q(s,a)$ has already learned about similar states. To enhance this behavior, the work \cite{schaul2015prioritized} proposes a Prioritized Experience Replay (PER) method that emphasizes the learning of those experiences that have bigger TD error in terms of Eq. \eqref{eq:loss}. The PER method imposes a probability $P_i$ of sampling an experience $i$ in the buffer proportional to its TD error
 
 \begin{equation}
    P_i = \frac{TD_i^\alpha}{\left(\sum_{\forall E \in Memory} TD \right) ^ \alpha}
    \label{eq:priority}
 \end{equation}
 
 with $\alpha$ being a parameter of how uniform the sampling is. To avoid the bias generated by the prioritized selection of bigger TD values, the loss generated by these experiences is weighted using Eq. \eqref{eq:importance_sampling}. In this importance-based sampling, $\beta \in [0,1]$ represents the level of compensation, and in this work it is annealed from a baseline value, to 1, as the learning progresses.
 
  \begin{equation}
   w_{i}=\left(\frac{1}{N} \cdot \frac{1}{P_i}\right)^{\beta}
    \label{eq:importance_sampling}
 \end{equation}

 \subsubsection{Censoring Q-values}

 As learning simultaneously to deal with the entropy minimization task and avoiding invalid actions is an arduous task, a simple modification of the classical DQL algorithm can be implemented to overcome this situation. The navigation map can be used to obtain the actions that will cause a collision in a given state $s$. Once those invalid actions are calculated, a censoring function $\eta(s,a) \in \mathbb{R}^{|A|}$ that represents whether an angle of $\Psi$ can be performed or not. Once the invalid actions are computed, the state $s$ is processed by the CNN and the final censored Q values $\hat Q$ are obtained. The agent can only choose those actions $a$ that are not marked as invalid by $\eta$, as depicted in Eq. \eqref{eq:censoring}.
 
  \begin{equation}
  \begin{array}{cc}
       \eta(s,a) = \begin{cases}
            1, \text{if $a$ is valid} \\
            -\infty, \text{if $a$ not valid}
           \end{cases}\\
           \; \\
       \hat Q(s,a) \leftarrow \eta(s,a) \circ Q(s,a) 
  \end{array}
    \label{eq:censoring}
 \end{equation}
 
 This technique allows the agent to choose the best actions learned in $Q(s,a)$ by greedily acting with basic knowledge of the scenario. This also simplifies the learning and reward shaping by eliminating the need for collision-related parameters, such as penalties and terminal conditions. The final algorithm can be seen in Algorithm \ref{alg:censoring}. The issue of reward shaping is discussed in \cite{Yanes2020} and involves not only the definition of penalization, but also the terminal condition of the episode. This condition determines how many collisions are permitted before the episode ends. If the memory buffer is filled with many collision experiences, the agent will overfit and will not learn efficient policies. On the other hand, if not enough collisions are experimented with, there is a high probability of collision, as there is a low probability of sampling collisions from the PER. This method allows all these aspects to be eliminated from the problem by means of a simple hard censoring approach. If navigable contours are available, the most efficient approach is not to learn those dynamics that are available all the time, which is the case for navigation boundaries.
 
\begin{algorithm}
\caption{Noisy Censoring-DQL}\label{alg:cap}
\begin{algorithmic}
\Require $Env, \alpha, \beta, \gamma, l_R, Memory, \varpi$
\Ensure $Q(s,a; \theta, \epsilon)$
\State Initialize $Q(s,a; \theta', \epsilon), Q^*(s,a; \theta'', \epsilon)$

\While{$episode \leq$ Learning budget} 
    \State Reset $Env$
    \State $s_t \gets s_0$ from $Env$
    \While{not done}
        
        \State $\epsilon \gets \epsilon'$ \Comment{Sample new noisy weights.}
        \State $\hat Q(s_t,a) \gets \eta(s_t,a) \circ Q(s_t,a)$ \Comment{Apply safe censoring.}
        \State $a_t \gets \argmax_{a} \hat Q(s_t,a; \theta, \epsilon)$
        \State $s_{t+1}, r_{t} = Env.step(a_t)$
        \State Store $<s_{t}, a_{t}, r_{t}, s_{t+1}>$ in $Memory$
        \State Update $Memory$ priorities using Eq. \eqref{eq:priority}.
        
        \State $\mathcal{B} \gets$ $ExpRep.sample$ \Comment{Prioritized sample.}
        \State $\epsilon \gets \epsilon''$ \Comment{Sample another noisy weights.}
        \State $\textbf{y} \gets r_t^\mathcal{B} + \gamma \max_{a'} \eta(s_t^\mathcal{B},a') \circ Q^*(s_{t+1}^\mathcal{B},a'; \theta', \epsilon)$
        \State $\mathcal{L} =  w^{\mathcal{B}} \times \left[ Q(s_t^\mathcal{B}, a_t^\mathcal{B}; \theta, \epsilon) - \textbf{y} \right]^2$
        \State $\theta \gets \theta + l_R \times \frac{\partial \mathcal{L}}{\partial \theta}$ \Comment{SGD step.}
        
        \State $\theta' \gets  (1-\varpi) \times \theta' + \varpi \times \theta$
        
        \If{$Env.distance \geq $ distance budget}
            \State $done \gets True$
        \EndIf
            
    \EndWhile
    
    \State $episode \gets episode + 1$
    
\EndWhile

\end{algorithmic}
\label{alg:censoring}
\end{algorithm}

\subsubsection{Reward function}

\changes{The design of an adequate reward function is fundamental for any DRL application \cite{Yanes2020}. The reward function $r(s,a)$ will define how good an action $a$ is in a state $s$ from the perspective of the optimization objective. In this work, the final optimization goal is to reduce the entropy. As the entropy is proportional to $|\Condicionada|$, the reward function must evaluate the actions that decrease this determinant by defining an information gain measure. According to \cite{SimRoy}, the most effective way to reduce the determinant follows the A-optimal criterion, which is to minimize the average of the eigenvalues $\lambda$ of $\Condicionada$. This will lead to a definition of the information value $I$ as seen in Eq. \eqref{eq:information}. This information measure has also been tested in other relevant works such as \cite{PopovicIPP}. The A-optimal criterion consists of reducing the diagonal values of $|\Condicionada|$ by taking measurements following Eq. \eqref{eq:conditioned}. This will be equivalent to weighting every action in terms of how much the action reduced the mean uncertainty in the monitoring domain $X$.}

\begin{equation}
    I = \sum_{i=1}^{n} \lambda_i = tr(\Sigma[X|X_{meas}])
    \label{eq:information}
\end{equation}

In the end, the information gain $\Delta I$ can be defined as the decrease in information when incorporating the next measurement. 

\begin{equation}
    \Delta I_{t+1|t} = I_{t} - I_{t+1}
    \label{eq:information_2}
\end{equation}

If an action has less information than $\Delta I_{min}$, a useless movement is considered and a penalization of $\kappa < 0$ is applied. When collisions are considered, an action that aims to move the vehicle to an unvisitable zone, a penalty of $c$. The final reward function $r(s,a)$ is represented in Eq. \eqref{eq:reward}.

\begin{equation}
r(s,a) = 
\begin{cases}
c, \quad \text{if $ (s,a) \rightarrow$ collision}\\
\kappa, \quad \text{if } \Delta I_{t+1|t} < \Delta I_{min}\\
\Delta I_{t+1|t}, \quad \text{otherwise}
\end{cases}
\label{eq:reward}
\end{equation}

\section{Results and Discussions}

This section describes the simulation settings, the performance metrics related to the monitoring task, the results of such simulations, and a comparative study between our proposed work and previous algorithms and heuristics.

\subsection{Environment setting}

\begin{figure}[t]
\centering
\includegraphics[width=\linewidth]{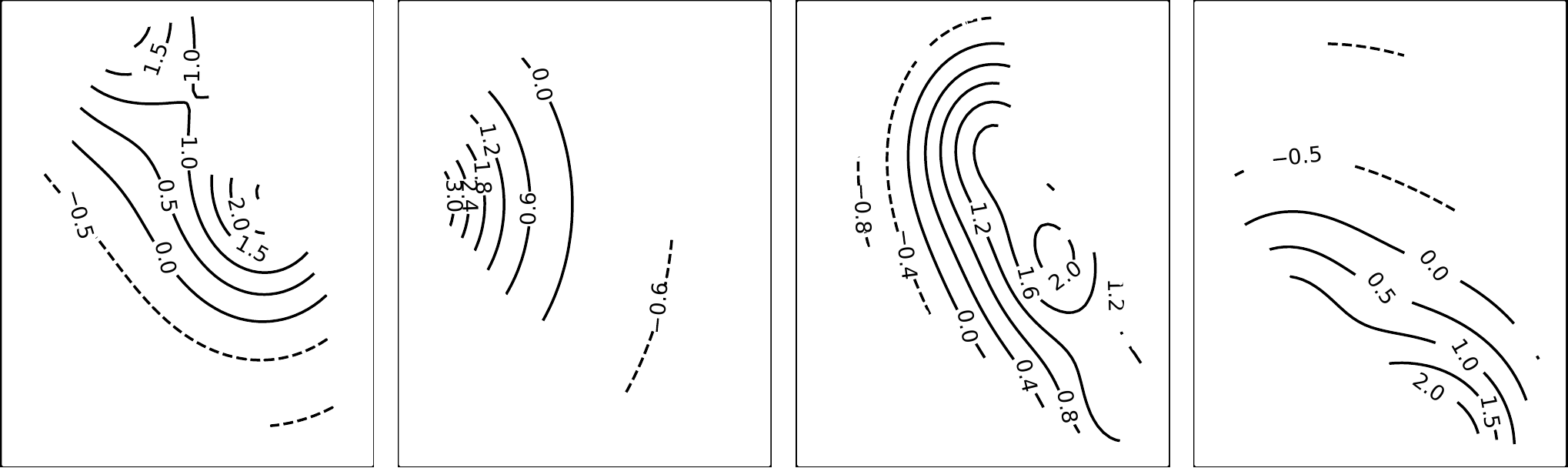}
\caption{Four different ground truths generated using a randomized Shekel function. The maxima of these ground truths can move with a constant speed of $v$ in random directions in the dynamic case.}
\label{fig:benchmarks}
\end{figure}

The environment used for training will be the lake Ypacaraí (Paraguay, $60km^2$). The distance budget for static missions is $45km$ and, in the dynamic case. In the temporal case, since we are concerned with continuous patrolling of the waters, the total distance will be $112,5 km$. This increase is necessary to compel with efficient single agent patrolling under the Ypacaraí water dynamics. In relation to WQ parameters, it is imposed that their behavior responds to a randomized benchmark function (Shekel function) previously used for monitoring tasks \cite{peralta2021bayesian,luis2021multiagent}. These functions represent a plausible characterization of real water resources such as the Mar Menor in Murcia\footnote{\url{https://deap.readthedocs.io/en/master/api/benchmarks.}}. We distinguish between the static case, where the benchmark functions do not change, and the dynamic case, where its maxima move with a random Brownian motion with a maximum speed of $v_{max}$ (see Figure \ref{fig:benchmarks}).

\subsection{Performance metrics}

To measure the performance of the method, several metrics are used. These metrics are considered to take into account different aspects of the monitoring: the WQ parameter estimation, outlier detection, and covered area. These metrics will be used in the comparison between algorithms. 

\begin{itemize}
    \item $I_t$: The non-gathered information, still available at time $t$. In the static case, the indicative metric of the performance is the information at the end of a mission $I_T$, and in the temporal case, as the $I_t$ grows in the zones further from the actual position of the ASV, it is considered the average non-gathered information once the initial 45$km$ are traveled. The lower this metric, the less useful information remains non-gathered in the scenario.
    
    \item $\boldsymbol{A}^{I}$: Represents the useful information area in $(km^2)$ covered by the ASV. A zone $x \in \mathbb{R}^2$ is considered covered if the uncertainty $\sigma(x)$ is less than 0.05. The higher the value of $\boldsymbol{A}^{I}$, the more thorough the coverage.
    
    \item $\boldsymbol{MSE}$: Mean square error between a regression model and the ground truth of the WQ variables. Two regression methods have been tested: GP as in \cite{peralta2021} and Space Vector Regressor (SVR) \cite{SVR}. Both regression methods use the same kernel in Eq. \eqref{eq:RBF}. For the temporal case, the sample time $t_{meas}$ is imposed as a variable like in \cite{DynGPBranke}. The lower the MSE, the better the environmental model is obtained.
    
    \item$\boldsymbol \xi$: Peak detection rate. In the presence of $k$ random peaks of algae blooms or contamination, the average rate of detected peaks is defined as $\xi = \mathbb{E}[k_{detected}/k]$. A local maximum is considered detected when the uncertainty $\sigma(x)$ at its location $x \in \mathbb{R}^2$ is less than 0.05. For the temporal case, to compel the dynamic scenario, the contamination agents can move through the scenario with speed, as explained in Section 5.1. A higher value of $\boldsymbol \xi$ indicates better detection of outliers and peaks.
    
\end{itemize}

\subsection{Learning settings}

All simulations were run using PyTorch on an Ubuntu 20.04 server with an RTX A600 GPU (42 Gb VRAM), 192Gb RAM, and two Intel Xeon Gold 5220R 2.20GHz. \changes{In both static and dynamic scenarios, the agent has been trained for $1 \times 10^4$ episodes. Note that in the static case, at most and in the absence of collisions, the vehicle will take 67 water quality samples, considering the value of $d_{meas} = 0.675 km$. In the case of the dynamic scenario, the episode will last 168 steps at most. Considering Algorithm 1, this implies a higher number of updates of the neural network weights. This is in line with the need for more training for the second problem, which is more complex}. 

\changes{The number of episodes has been chosen using a similar order to that of previous similar applications \cite{Yanes2020}. It has been observed that the number of episodes is sufficient so that, in both cases, the policy converges to a local optimum with a sufficiently adequate execution time. Regarding training times, it is important to note that the use of GPUs is highly recommended. In the static case, the simulation has a training time of approximately 4.5 hours for the static case and 11.3 hours for the dynamic case, which has a longer duration of episodes.}

The SGD optimizer is Adam with a learning rate $l_R$ of $1 \times 10^{-4}$. \changes{To compare the $\epsilon$-greedy exploration policy with the intrinsic exploration of the noisy neural network, separate experiments have been performed under the same conditions. In the case of the $epsilon$-greedy policy, the value of $\epsilon$ has been imposed to decay within the interval $[\epsilon_{max}, \epsilon_{min}] = [1, 0.05]$ over the training time. Then, $\epsilon$ decays from the very beginning of the training to 3000 episodes. From this point, if the exploration phase is extended, no improvement has been observed. In any case, when evaluating the resulting policies, it is imposed a full-greedy action selection without any exploration $\epsilon = 0$. All other hyperparameters and other simulation settings are summarized in Table \ref{tab:hyperparam}.}

\begin{table}[]
\centering

\begin{tabular}{p{0.45\linewidth}p{0.45\linewidth}}
\hline
\textbf{Learning hyperparameter}       & \textbf{Value}      \\ \hline

Learning rate                                & $1  \times 10^{-4}$ \\
Target update constant $\varpi$ & $1 \times 10^{-4}$ \\
Batch size $|\mathcal{B}|$                                & 64                  \\
Unbias $\beta$ interval                         & $[0.5, 1]$          \\
Priority value $\alpha$                            & 0.5                 \\
Discount factor $\gamma$                          & 0.99                \\ 
$\epsilon$ interval                                        & [1, 0.05]           \\
Collision penalty  $c$                      & -1                  \\
Redundancy penalty $\kappa$                & -0.5                \\
Information threshold $\Delta I_{min}$    & 0.01                \\
\hline
\textbf{Environment parameter}          & \textbf{Value}      \\ 
\hline
$d_{meas}$                                              & 0.675 $km$   \\
RBF lengthscale $l$                          & 1.125 $km$   \\
Forgetting factor $\tau$                       & 0.03         \\
Max. peaks speed $v_{max}$             & 0.3   $m \cdot s^{-1}$ \\

\bottomrule
\end{tabular}%

\caption{\changes{List of environment parameters, simulation settings, and hyperparameters related to the proposed DQL algorithm.}}
\label{tab:hyperparam}
\end{table}

\subsection{Learning results}

\begin{figure}[t]
\centering
\includegraphics[width=\linewidth]{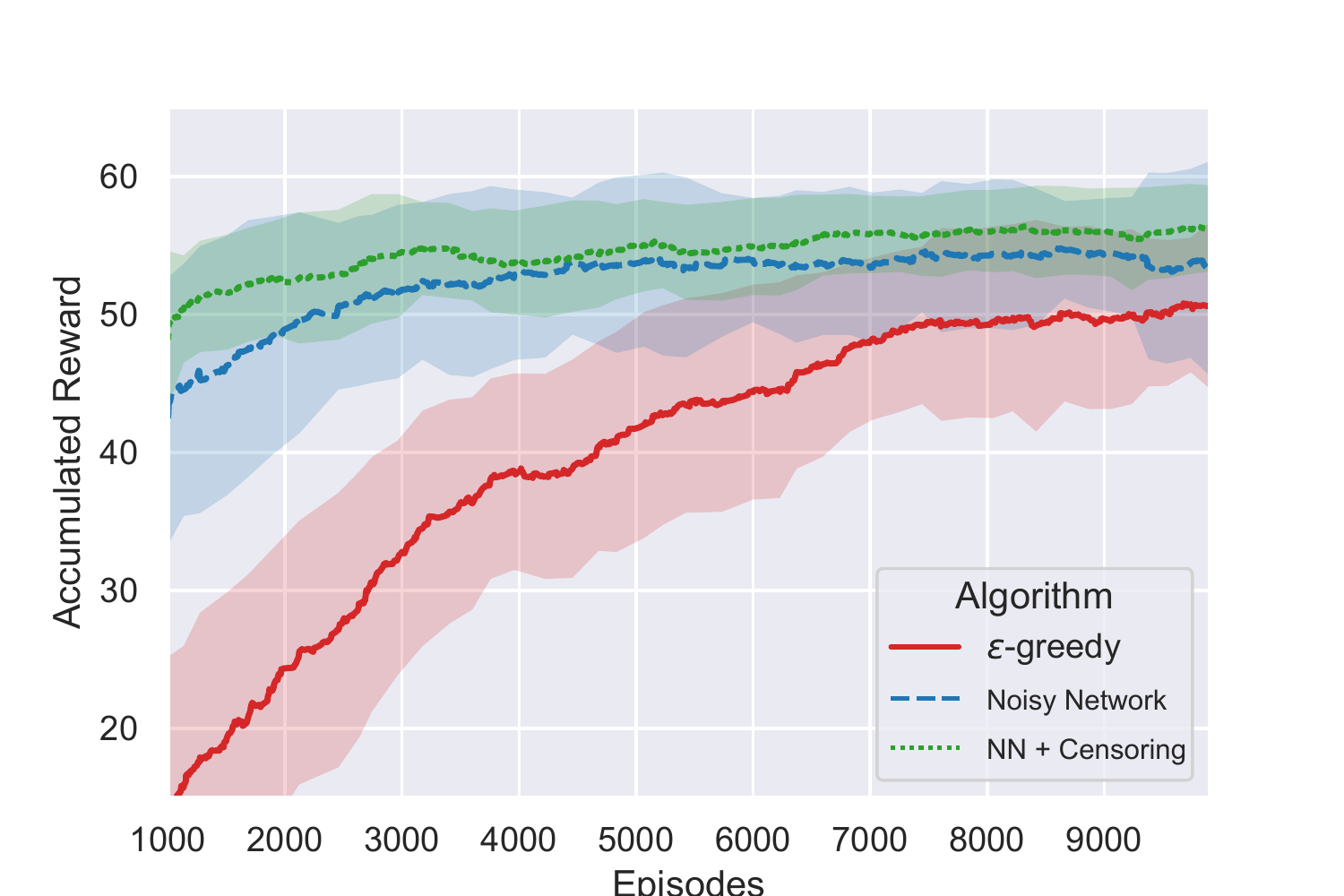}
\caption{Accumulated reward over training time in the static scenario. It represents the average reward with a moving window of 100 episodes with its standard deviation. Note that Q-Censoring has collision-free training and no penalization is applied.}
\label{fig:static}
\end{figure}

\begin{figure}[t]
\centering
\includegraphics[width=\linewidth]{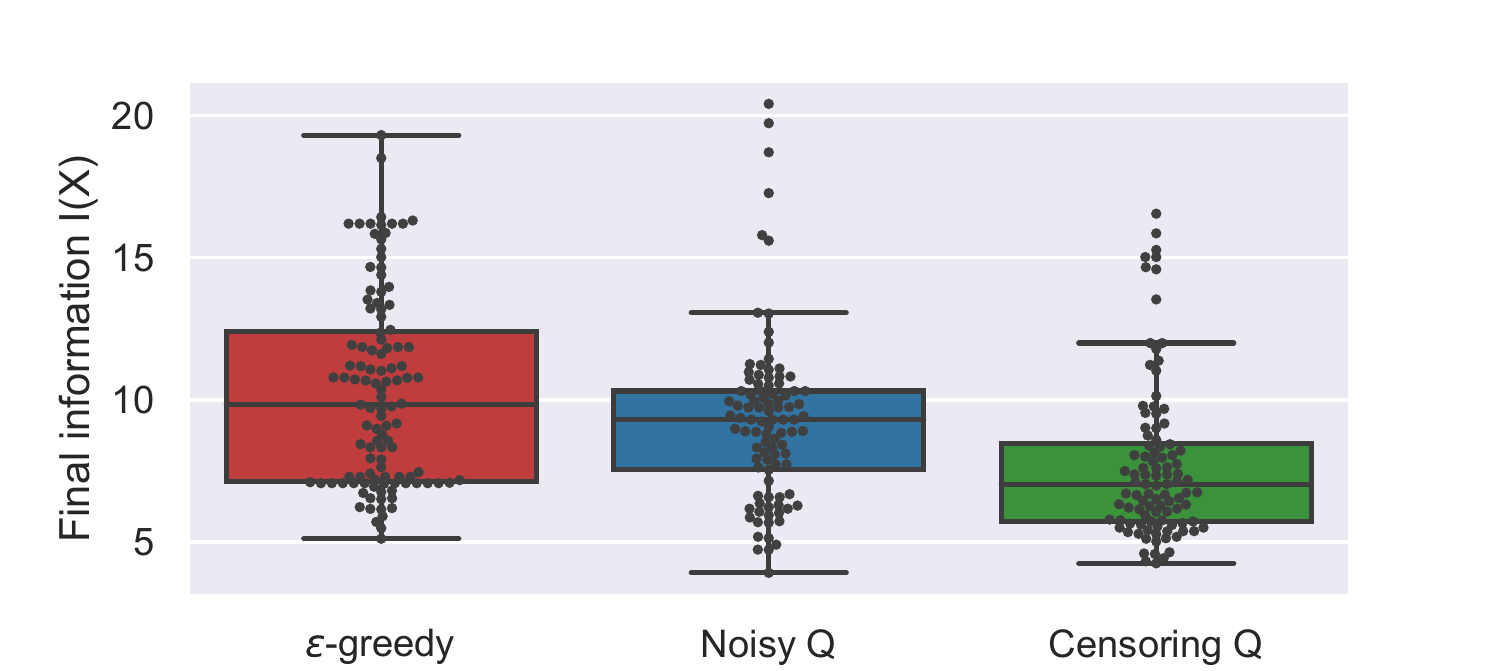}
\caption{Box plot of information at the end of the mission, for 100 different starting points, in the static case.}
\label{fig:box_plot_entropy_comparison_static}
\end{figure}

\begin{figure}[t]
\centering
\includegraphics[width=\linewidth]{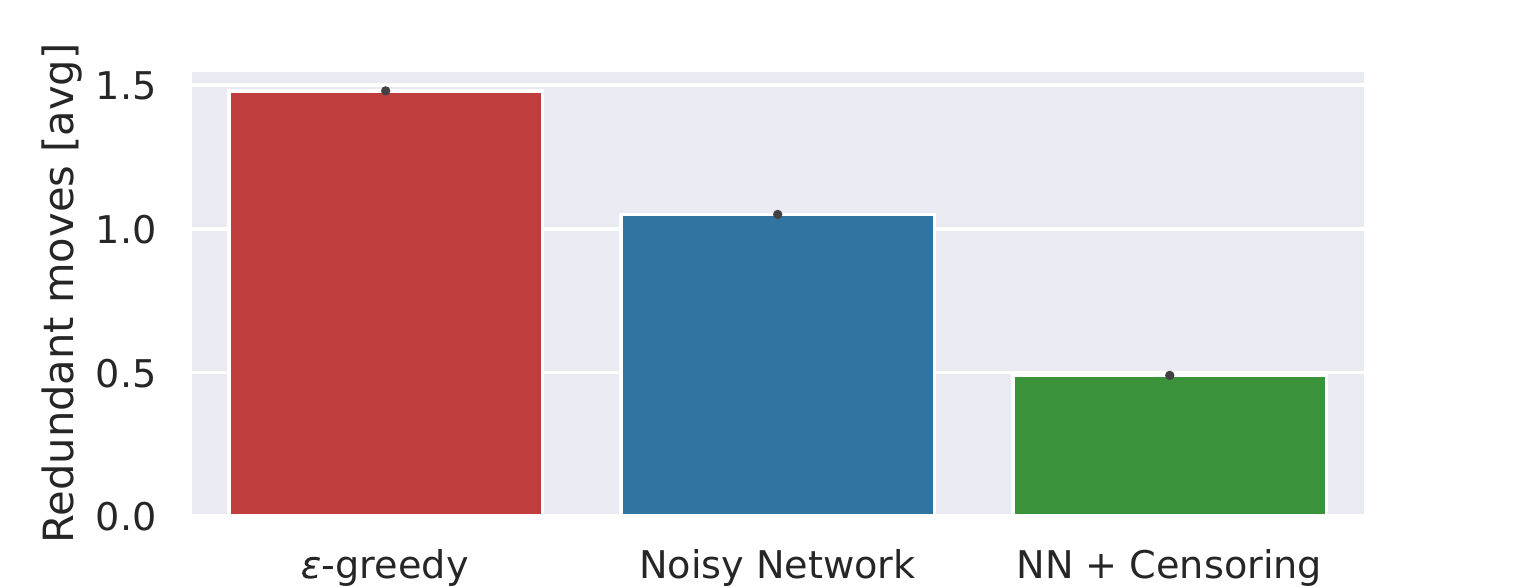}
\caption{Number of useless moves on average, where it is applied a penalization of $\kappa$ for the three variants of the DRL approach.}
\label{fig:redundant_moves}
\end{figure}

The learning results can be seen in Figures \ref{fig:static} for the static case. The advantage of using a noisy network can be seen in terms of convergence and training speed. In this first scenario, the use of the noisy network will increase the learning efficiency and, at the same time, eliminate the need to design the $\epsilon$-greedy strategy and its hyperparameters. The noisy network converges to an explorative-explotative balanced policy, which can be seen as incorporating the exploration behavior into the policy. The Noisy + Q-Censoring strategy, in the static case, slightly overcomes the average reward by approximately 30\% with respect to the $\epsilon$-greedy counterpart and by 20\% with respect to the only Noisy Network case. Although the improvement is not overly significant, the Q-Censoring strategy translates its advantage in the fact that the convergence is reached earlier and, once again, it eliminates the need of designing the collision penalty. In \cite{Yanes2020}, it was studied how the penalty affects convergence and how it can cause poorer performance under certain conditions. In Figure \ref{fig:box_plot_entropy_comparison_static}, the entropy is shown at the end of an episode after evaluating the proposed method 100 times starting from random points on the map at each run. The noisy policies return more informative paths (12\% improvement), and the noncensored version still provides an effective policy to deal with collisions at the same time the entropy is reduced. It is also noticeable that the number of average redundant/useless movements that produce $I^t < I_{min}$ (see the reward function Eq. \eqref{eq:reward}) is 4 times higher in the $\epsilon$-greedy strategy and 2.75 times higher in the noncensored version than in the censored version (see Figure \ref{fig:redundant_moves}). This indicates a better policy in terms of the information gathering capabilities of the proposed method. 

\begin{figure}[t]
\centering
\includegraphics[width=\linewidth]{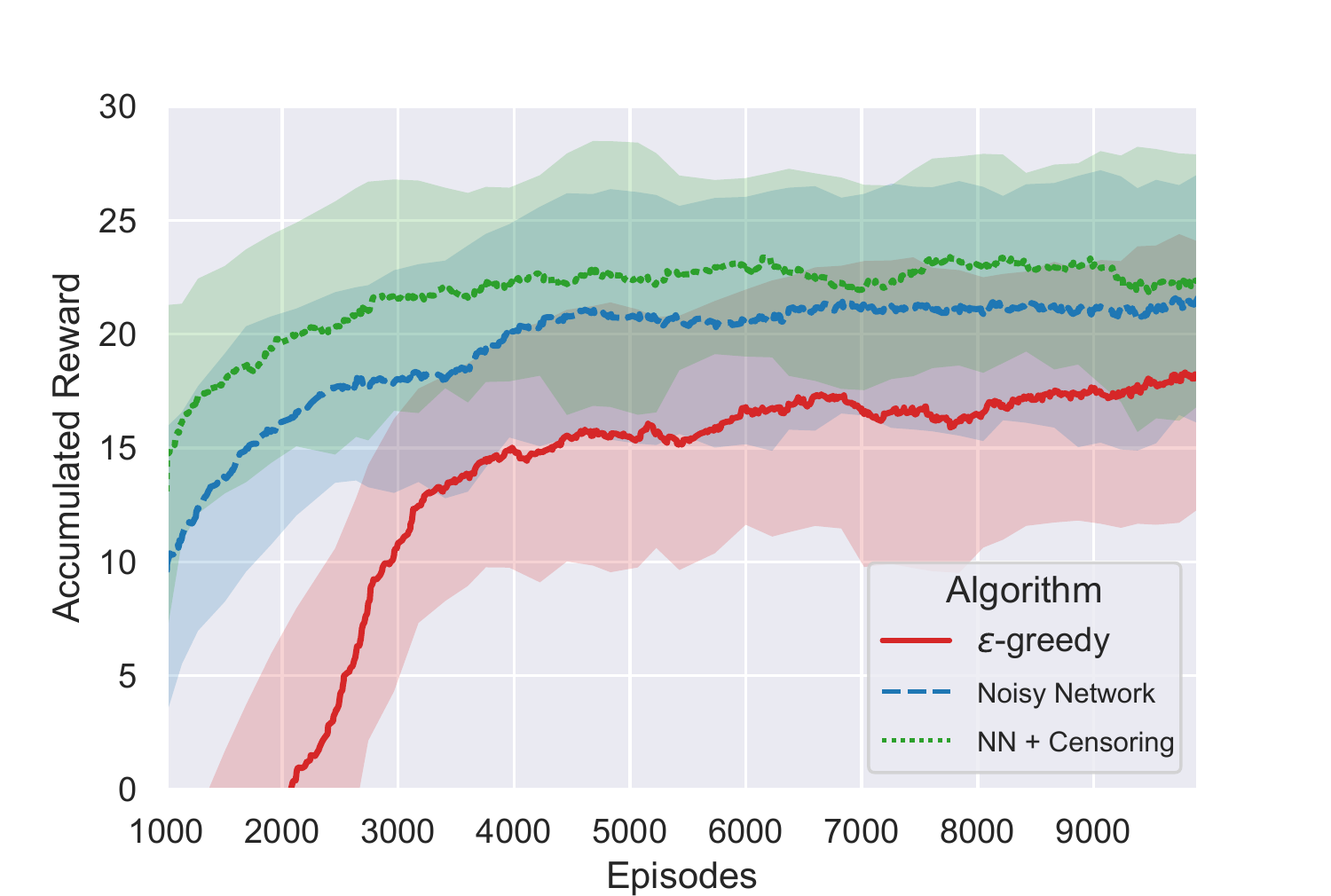}
\caption{Accumulated reward over training time in the dynamic scenario. Note that the total reward changes from the static case, as the amount of information, once the map has been covered after a while, is much lower than in the static scenario.}
\label{fig:dynamic}
\end{figure}

\begin{figure}[t]
\centering
\includegraphics[width=\linewidth]{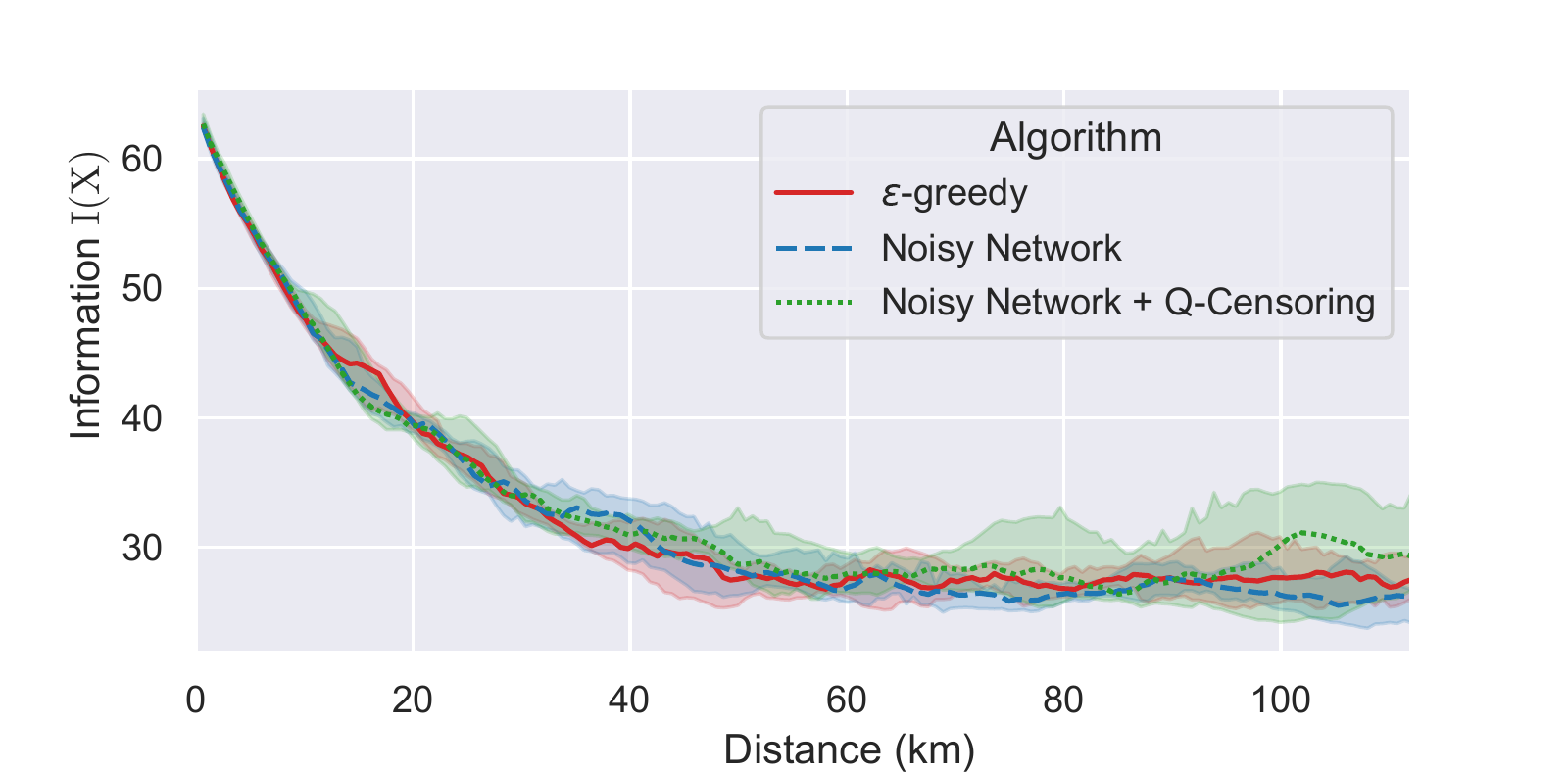}
\caption{Comparison of non-gathered information $I_t$ in the scenario for the three DRL algorithms. It represents the average of 100 missions with different starting points and its standard deviation.}
\label{fig:entropy_comparison_dynamic}
\end{figure}

Regarding the dynamic case, in Figure \ref{fig:dynamic} it can be seen how the Q-Censoring strategy overcomes the other methods. As the episode in this particular case is 2,5 times bigger, both the noisy and $\epsilon$-greedy strategies tend to generate collisions at a certain point. Using the Q-Censoring strategy, learning is simplified and allows for a higher reward in time with little effort. The Non-Censoring Noisy Network version can be seen as a curricular learning like in \cite{CurricularLearning}, where it is assimilated first through the boundaries of the map, and later, the entropy reduction policy is optimized. The censored strategy alleviates this condition, especially in longer episodes, where the possibility of collision is greater. In Figure \ref{fig:entropy_comparison_dynamic}, it shows the result of the nongathered information $I_t$ of 100 episodes using the three DRL approaches. Two important aspects must be remarked: I) As the uncertainty grows within time, it is not possible to gather information with near-zero values, as in the static case. It is only possible to find an equilibrium where the patrolling path reduces the information to a certain point and avoids unnecessary redundant movement. II), the three DRL methods result in valid temporal patrolling strategies, but the censoring version policy is obtained with 50\% less episodes and therefore is the best candidate for more complex scenarios.

\subsection{Comparison with other methods}

To evaluate the performance of the proposed method in IPP and IPPP, different algorithms have been tested in the Ypacaraí scenario. To fairly compare the ability of the other algorithms to reduce the available information, only safe actions can be performed, considering the navigation map:

\begin{itemize}
\item \textbf{Safe Random Agent}: \changes{Actions are randomly chosen from the action space $\textbf{A}$ if they are valid. This algorithm is useful for comparing the difficulty of the patrolling problem, in a way similar to \cite{arzamendia2019TSP}.}

\item \textbf{Lawn Mower}: The Lawn Mower (LM) algorithm is a strategy that generates intensive coverage paths. The LM agent chooses a random direction to start and goes straight until it is not possible to go further. Then, the agent returns in a parallel straight line, one step forward from the previous path.

\item \textbf{Non-Redundant Random Coverage}: \changes{The Non-Redundant Random Coverage (NRRC) algorithm randomly selects a direction of movement from $\textbf{A}$, different from the previously selected to avoid retracing its steps. Once the vehicle cannot advance in this direction, a new direction is sampled. This is a variation of the Intelligent Random algorithm used for comparisons in \cite{ViserasDeepIG}.}

\item \textbf{I-greedy Strategy}: The $I$-greedy strategy selects the highest point of uncertainty on the map and, using a local path planner, guides the vehicle to this point without collisions.
\end{itemize}

\begin{figure}[thb]
\centering
\includegraphics[width=\linewidth]{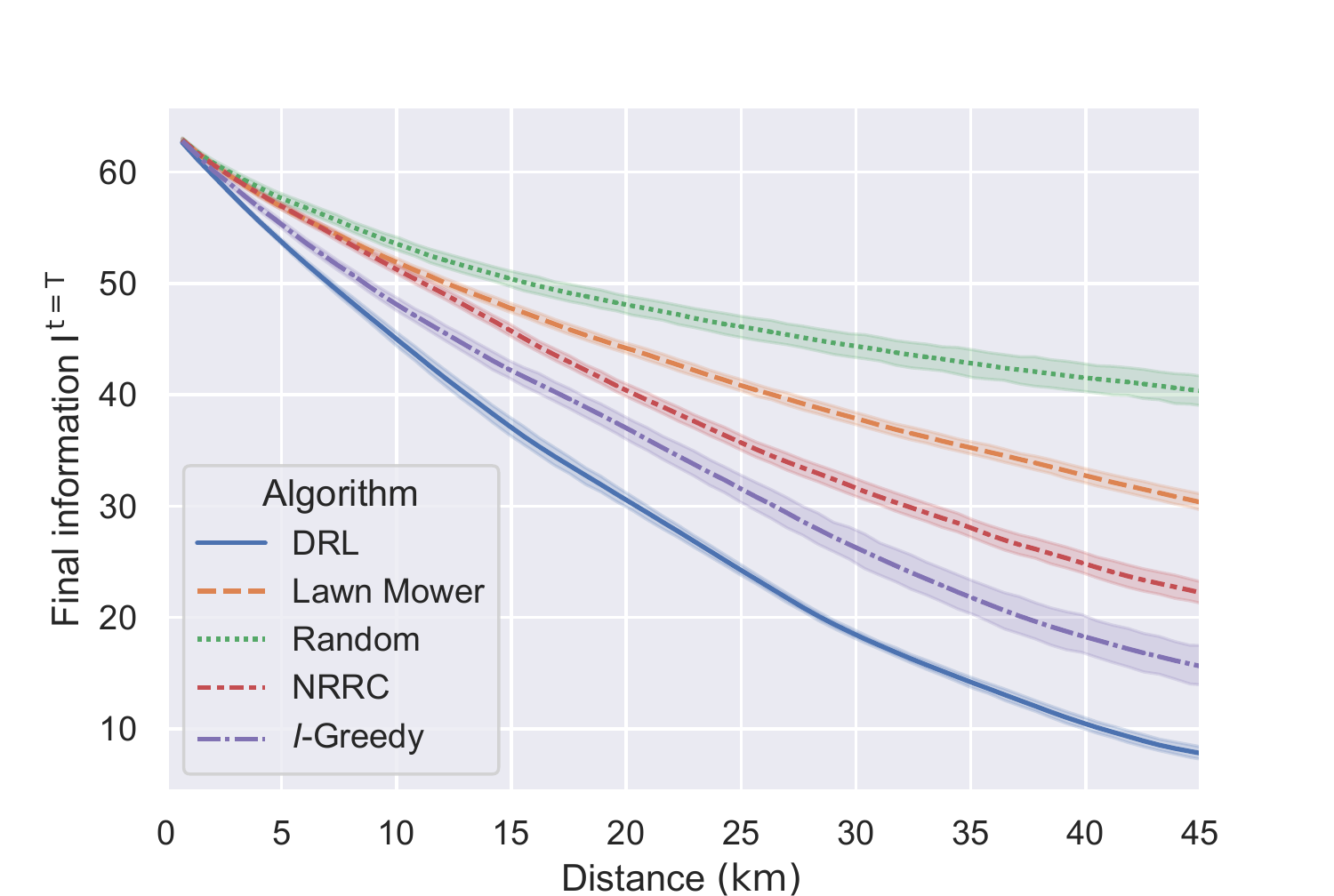}
\caption{Comparison of the average information in the static scenario for all algorithms after 100 missions from different starting points and its standard deviation.}
\label{fig:entropy_comparison_static}
\end{figure}

\begin{figure}[thb]
\centering
\includegraphics[width=\linewidth]{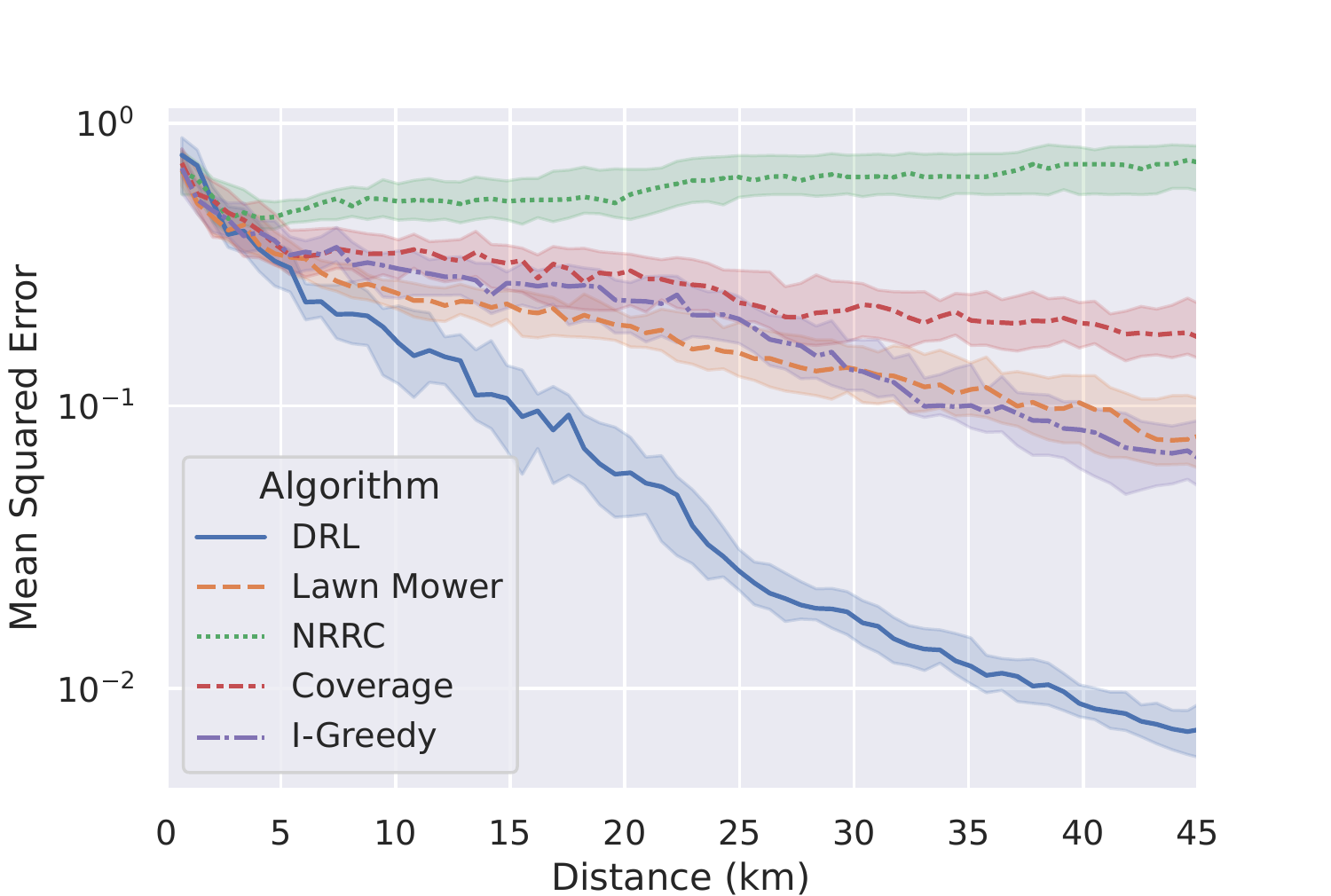}
\caption{Mean squared error using a GP with the proposed kernel after 100 missions (static case) with random ground truths for every algorithm. The average is represented within its standard deviation.}
\label{fig:MSE_comparison_static}
\end{figure}

The results of the static case are presented in Table \ref{tab:static_results}, for 100 different starting point routes. These results clearly show how the proposed method is able to obtain the most informative paths compared to the other algorithms. The information obtained at the end of the episode is 49\% lower than the second best algorithm ($I$-greedy). The DRL agent is able to gather more information in much less time, as seen in Figure \ref{fig:entropy_comparison_static}. The DRL policy is also less affected by the starting point in terms of the standard deviation of the information, which can be seen as the robustness of the informative trajectories. The proposed algorithm also obtains a policy capable of detecting 60\% randomly located contamination peaks in the environment, which is a 51\% improvement on average over the other algorithms. Improvement is also significant in terms of MSE using both regression methods. On the one hand, when using a GP, the MSE is reduced to 0.019, which represents an average error of 13.7\% across the water surface (see Figure \ref{fig:MSE_comparison_static}) On the other hand, the SVR returns an average MSE of 0.032, again an average 17,8\% error, lower than in any other acquisition strategy tested. 

It is noticeable that the acquisition path is relevant to the modeling problem and also to the regression method used. GPs obtained, on average, better solutions for the static case, independent of the path. Nevertheless, it is proven that the error in the model is conditioned on the information gathered, which is the reduction of the entropy. In this static case, the lower the entropy, the lower the error in the estimation for different regressors. This suggests that optimizing under this informative entropic criterion produces useful coverage paths for many purposes: model estimation, patrolling contamination peaks, complete coverage, etc. 

\begin{figure*}[ht]
\centering
\includegraphics[width=\linewidth]{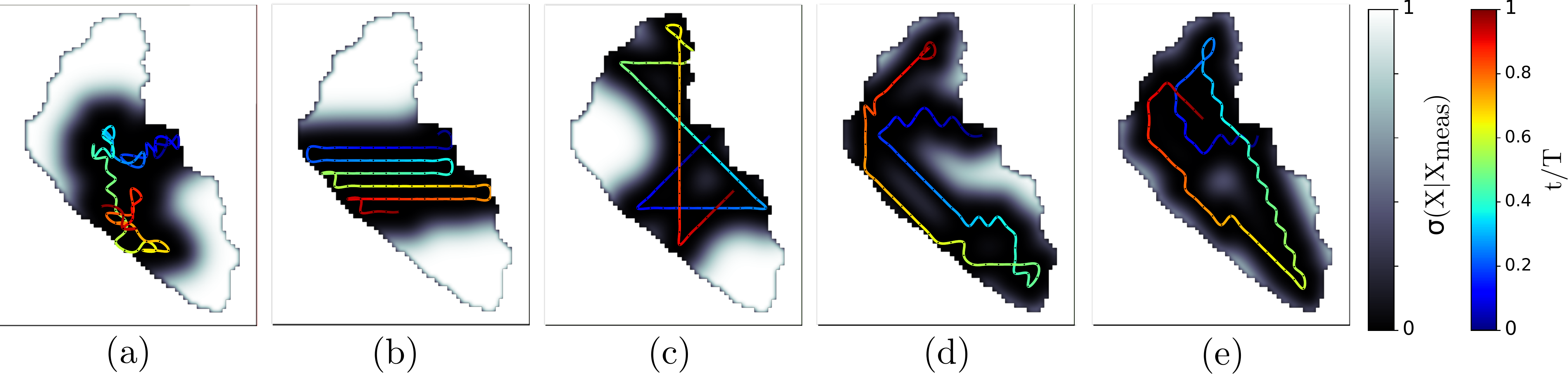}
\caption{Paths generated by common coverage path planning algorithms in the static scenario: (a) Random, (b) Lawn mower, (c) Nonredundant Random Coverage, (d) $I$-greedy, and (e) the proposed DRL approach. All paths start in this figure from the same deploy zone for comparison. The path is colored according to the time $t$ divided by the mission time $T$. The grayscale background color represents the uncertainty $\sigma(x)$ of a particular zone $x$.}
\label{fig:solutions_static}
\end{figure*}

Finally, if the final paths are studied as seen in Figure \ref{fig:solutions_static}, it is clear that they are less redundant in the proposed DRL policy. The paths generated by the LM approach tend to be too thorough and do not take into account the redundancy of the information in consecutive and near samples. In a different way, the greedy approach does not consider any further information than the maximum uncertainty position, which inevitably leads to redundant subpaths in the way of reaching the most informative point. The proposed algorithm takes into account not only the next point, but also the future discounted information available, and obtains, even in the final steps when there is little information left, significant measurements.

\begin{table*}[thb]
\centering
\renewcommand{\arraystretch}{1.5}
\begin{tabular*}{\linewidth}{@{\extracolsep{\fill}}cccccccccccc@{\extracolsep{\fill}}}
\hline
 &
  \multicolumn{2}{c}{\textbf{$\boldsymbol I_T$}} &
  \multicolumn{2}{c}{\textbf{$\boldsymbol A^r$}} &
  \multicolumn{2}{c}{\textbf{$\boldsymbol \xi_{rate}$}} &
  \multicolumn{2}{c}{\textbf{$\textbf{\text{MSE}}_{GP}$}} &
  \multicolumn{2}{c}{\textbf{$\textbf{\text{MSE}}_{SVR}$}} \\ \hline
\textbf{Algorithm} 
              & Mean                     & Std.   & Mean    & Std.  & Mean   & Std.   & Mean  & Std.  & Mean   & Std.  \\ \hline
DRL           & \textcolor{red}{7,722}   & 0,30   & \textcolor{red}{52,73}   & 3,52  & \textcolor{red}{0,602}  & 0,14  & \textcolor{red}{0,019} & 0,10 & \textcolor{red}{0,032}  & 0,03 & \\ \hline
Lawn Mower    & 30,486  & 3,81                    & 34,34          & 3,32  & 0,393  & 0,12  & 0,121 & 0,13 & 0,394  & 0,29 & \\ \hline
Random        & 40,302  & 6,77                    & 22,36          & 5,52  & 0,256  & 0,15  & 0,741 & 0,40 & 0,755  & 0,57 & \\ \hline
NRRC          & 22,128  & 5,29                    & 34,16          & 2,96  & 0,377  & 0,16  & 0,252 & 0,25 & 0,107  & 0,13 & \\ \hline
$I$-greedy    & \textbf{15,524} & 0,43            & \textbf{47,18}   & 7,87  & \textbf{0,541}  & 0,16  & \textbf{0,132} & 0,17 & \textbf{0,116 } & 0,19 & \\ \hline
\end{tabular*}
\caption{Metrics obtained by executing all algorithms after 100 episodes for the static case }
\label{tab:static_results}
\end{table*}
\renewcommand{\arraystretch}{1}

\begin{figure}[t]
\centering
\includegraphics[width=\linewidth]{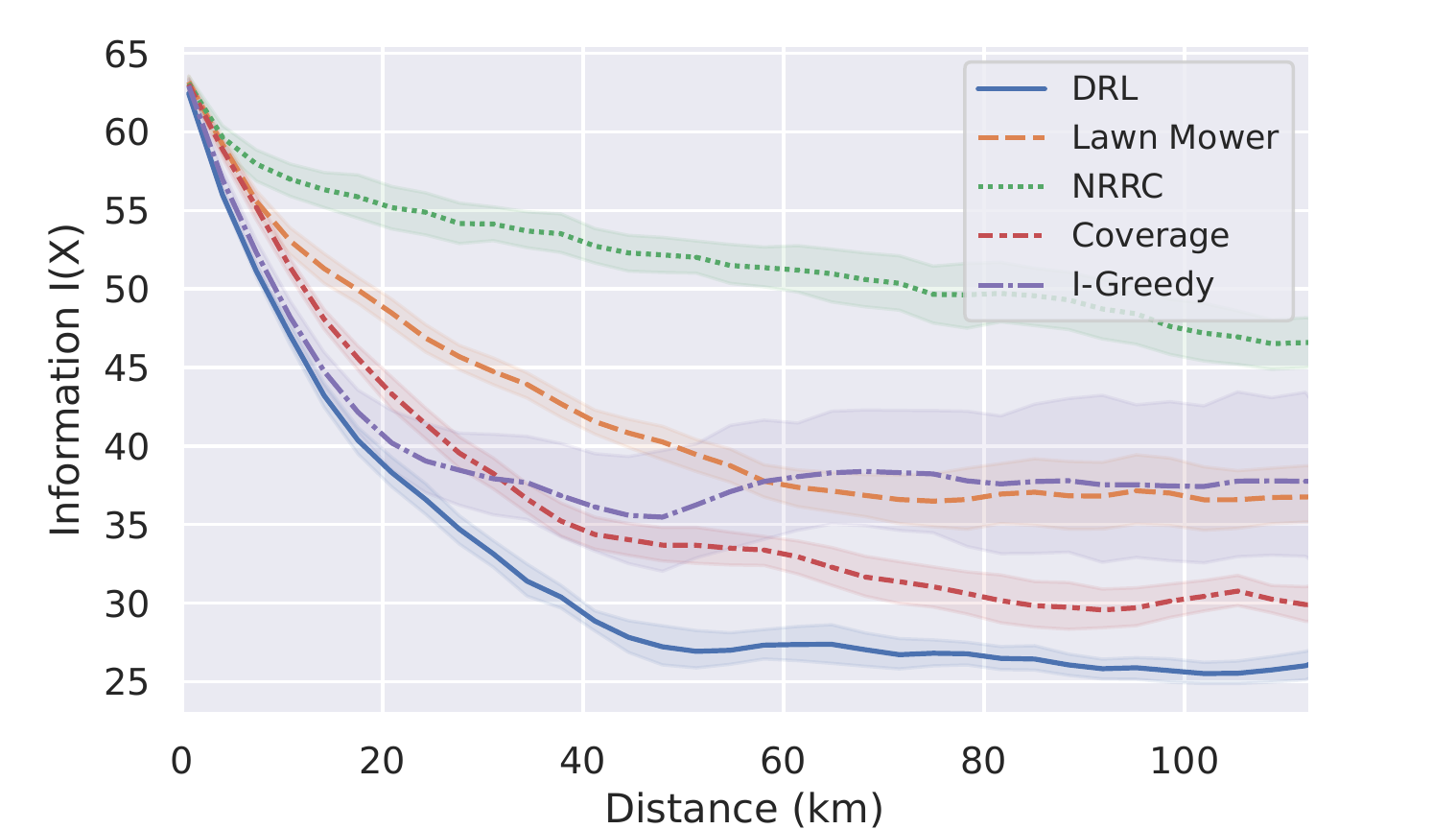}
\caption{Comparison of the average information in the temporal scenario for all algorithms after 100 missions from different starting points and its standard deviation.}
\label{fig:entropy_comparison_temporal}
\end{figure}

In the temporal scenario, the results can be interpreted in a similar way (see Table \ref{tab:temporal_results} for complete results). The proposed algorithm is able to obtain 30\% more informative paths on average than the other proposed heuristics. The DRL policy produces informative patrolling paths that can reduce entropy and maintain stability as information from the unvisited zone regenerates (see Figure \ref{fig:entropy_comparison_temporal}). When analyzing the ability of the algorithm to detect random peaks, the proposed DRL approach obtains similar but slightly better results than the second-best algorithm (LM). The LM algorithm, as it produces very extensive coverage paths, is able to detect those moving peaks, but at the expense of the surrogate model accuracy, which is far from the one obtained using the DRL method (a 51\% worse contamination model).

\begin{figure}[t]
\centering
\includegraphics[width=\linewidth]{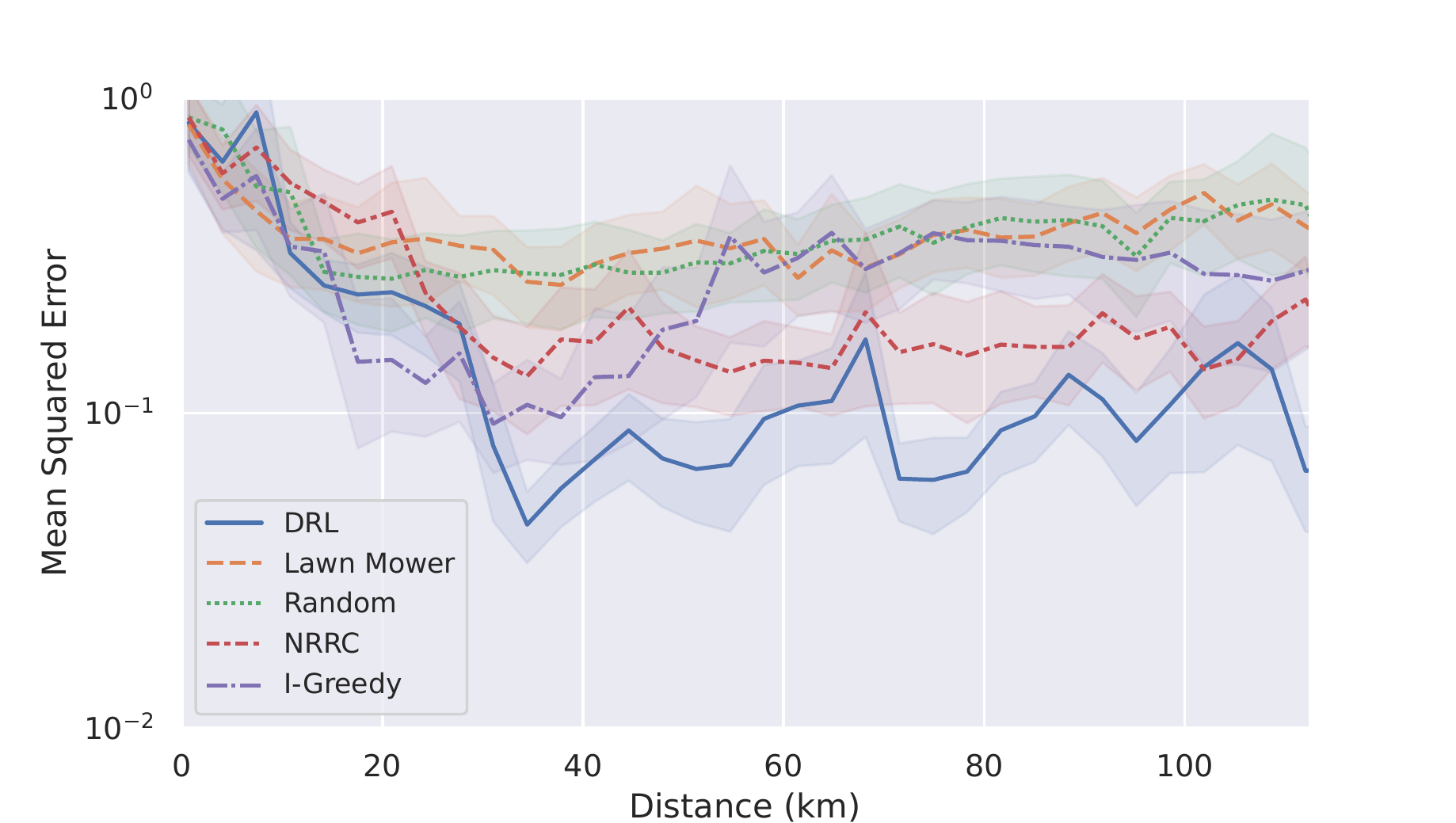}
\caption{Mean squared error using a GP with the proposed kernel after 20 missions starting from the same point (temporal case) with random ground truths. The average is represented within its standard deviation.}
\label{fig:MSE_comparison_TEMPORAL}
\end{figure}

Both regression methods used in the temporal case are modified to consider the acquisition time of a sample as a measured parameter, as explained above (see \eqref{eq:temporal_entropy}). The RBF kernels are consequently modified to include a time-dependent third dimension. It was also considered an estimation horizon of 67 samples in the past (the final number of samples in the static scenario), so the very first measurements were discarded for estimation, as they cannot be assumed valid anymore. With this simple temporal regression method, both GPR and SVR are able to converge in temporal-dependent models of the WQ variables with MSE values of 0,136 and 0,101, respectively, for each regressor. These values are obtained on average once the ASV has completed the first 45$km$ (the distance limit in the first static scenario). As can be induced from Table \ref{tab:temporal_results}, the MSE values are higher than in the static case, as the benchmark function changes over time. The model can only be adjusted when the old zones are revisited and the regressor updates its state information. In a further test, after 20 missions starting from a single starting point, the resulting MSE using GP is not only lower on average, but the lower bound is reached earlier in the DRL proposal (see Figure \ref{fig:MSE_comparison_TEMPORAL}). In the SVR, it is worth mentioning that not all policies result in a convergence of the surrogate model. In the case of the Random Agent, the NRRC, and the greedy approach, the measurements do not conform to an adequate set to converge over time. However, when converging, as happens in the DRL approach, the error can be even lower than in the GPR counterpart.

\begin{figure*}[ht]
\centering
\includegraphics[width=\linewidth]{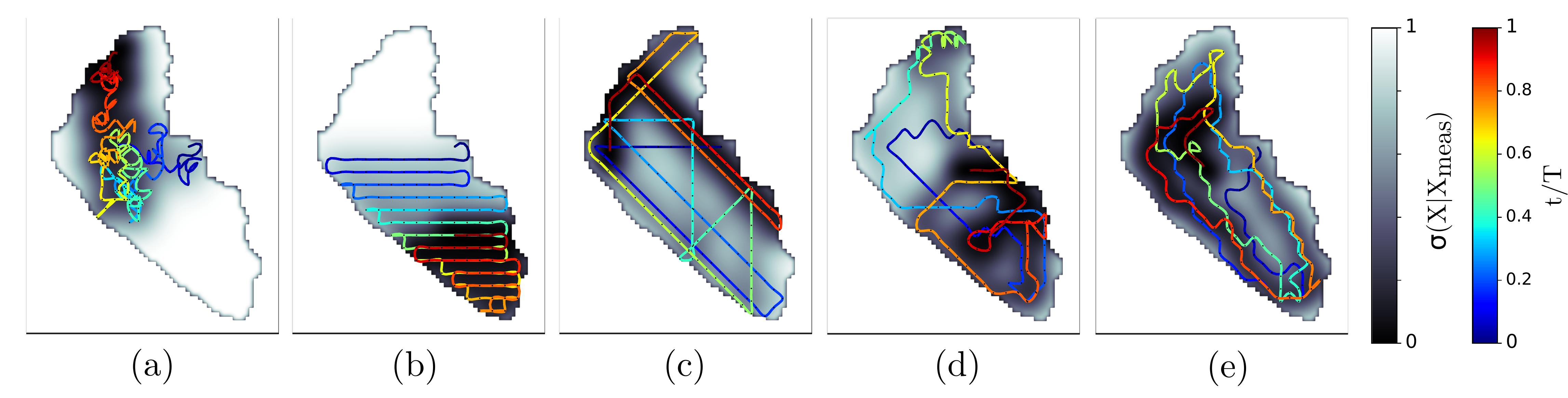}
\caption{Paths generated by common coverage path planning algorithms in the temporal scenario: (a) random, (b) lawn mower, (c) nonredundant random coverage, (d) $I$ -greedy, and (e) the proposed DRL approach. The path is colored according to the time $t$ divided by the mission time $T$. The gray background color represents the uncertainty $\sigma(x)$ of a particular zone $x$.}
\label{fig:solutions_temporal}
\end{figure*}

The example paths in Figure \ref{fig:solutions_temporal} show how the DRL policy tends to generate cyclic patrol routes to keep the growing uncertainty low. LM algorithm, while intensive in the final steps, tends to generate high redundancy coverage. The greedy approach, as the maximum uncertainty is always growing, tends to oscillate and fails the patrolling task.  The NRRC is the only algorithm that can compete in terms of the MSE model but easily generates long redundant paths.  

\begin{table*}[t]
\centering
\renewcommand{\arraystretch}{1.5}
\begin{tabular*}{\linewidth}{@{\extracolsep{\fill}}cccccccccccc@{\extracolsep{\fill}}}
\hline
 &
  \multicolumn{2}{c}{\textbf{$\boldsymbol I_T$}} &
  \multicolumn{2}{c}{\textbf{$\boldsymbol A^r$}} &
  \multicolumn{2}{c}{\textbf{$\boldsymbol \xi_{rate}$}} &
  \multicolumn{2}{c}{\textbf{$\textbf{\text{MSE}}_{GP}$}} &
  \multicolumn{2}{c}{\textbf{$\textbf{\text{MSE}}_{SVR}$}} \\ \hline
\textbf{Algorithm} 
              & Mean    & Std.    & Mean    & Std.  & Mean   & Std.   & Mean  & Std.  & Mean   & Std.  \\ \hline
DRL           & \red{26.673}  & 4.75    & \textbf{809.5}   & 32.4  & \red{0.725}  & 0,10  & \red{0.136} & 0,20    & \red{0.101}  & 0,07 & \\ \hline
Lawn Mower    & 36.765        & 4.15    & \red{913.7}   & 96.0  & \textbf{0,700}  & 0,20  & 0.342 & 0,22   & $>$1   & $>$1   & \\ \hline
NRRC          & 46.585        & 3.63    & 811.9   & 34.5  & 0.325  & 0,14  & 0.449 & 0,42   & $>$1   & $>$1   & \\ \hline
Coverage      & \textbf{29.904}  & 2.52    & 673.2   & 36.6  & 0.565  & 0,16  & \textbf{0,184} & 0,20   & \textbf{0,417}  & 0,20 & \\ \hline
$I$-greedy    & 37.755           & 11.77   & 587.1   & 163.9 & 0.525  & 0,20  & 0,471 & 0,65   & $>$1   & $>$1 & \\ \hline
\end{tabular*}
\caption{Metrics obtained by executing all algorithms after 100 episodes for the temporal case. Those MSE values with $>1$ correspond to those cases in which the regression method fails to converge and returns a useless model with errors greater than 1.}
\label{tab:temporal_results}
\end{table*}
\renewcommand{\arraystretch}{1}

\subsection{Discussion of the results}

The proposed algorithm has been proven to be effective in both IPP and IPPP applications in terms of monitoring metrics and learning efficiency. It is important to highlight the following aspects of the results:

\begin{itemize}

\item The use of noisy neural networks allows for efficient exploration of the action-state domain, resulting in an improvement in the information collected on the routes (20\% improvement). Additionally, the design of the $epsilon$-greedy strategy is no longer necessary.

\item The use of Q-censoring allows for better results than baselines (8\% improvement) in the static case and an increase in convergence speed (50\% faster).

\item The use of Q-censoring also allows the user to not impose any collision penalty, resulting in an improvement in policy performance in terms of information collected.

\item The agent in the static case produces significantly better monitoring results than the other heuristics (13.7\% on average among all algorithms in terms of MSE).

\item In the temporal case, the improvement is not only significant in terms of model error (8\% better than the best among the other algorithms), but also in other metrics, such as the detection rate of contamination agents (37\% more detections on average among the other algorithms).

\item On a qualitative level, the routes of both algorithms are less redundant and better understand the boundaries of the map and the information constraints on them.

\end{itemize}

\section{Conclusions and future lines}

\changes{The monitoring and patrolling task in big water resources requires an efficient path planning policy that gathers the most information available at the same time the navigation boundaries are considered. This work proposes a Deep Reinforcement Learning framework that optimizes a convolutional policy based on a visual state of the informative scenario using the entropy minimization as the surrogate objective for the patrolling task. Knowledge assimilation is enhanced using state-of-the-art methods such as Noisy Networks and Prioritized Experience Replay to dive into a more exploratory policy that can deal with two cases of monitoring: the static scenario and the temporal patrolling case. Ultimately, a Q-Censoring strategy has been proposed that permits scenario-gnostic learning to avoid dangerous movements towards the land. This approach has proven to be much more efficient than the classic $\epsilon$-greedy strategy and the noisy baseline, especially in the static case. The difference also lies in the convergence speed. As the Q-Censored approach does not deal with the collision problem, the speed of learning is significantly increased. The result of this framework will allow the establishment of efficient routes in the collection of water quality data and in many other applications, such as the measurement of radiation levels, gas sources, etc., without loss of generality. This framework is particularly useful due to the vehicle safety criterion and due to the formulation of information based exclusively on uncertainty.} 

\changes{The algorithm, in the end, results in a more useful policy than other approaches like the Lawn Mower or $I$-greedy, since it learns how to plan in an optimization horizon according to the tailored reward function. This improvement in entropy minimization is translated into lower model errors and higher detection rates of contamination peaks. However, regression methods have been shown to be very sensitive to acquisition policies and require further research. The convergence of the regression is not guaranteed, and the estimation error could be used as part of the reward function. Another future line is the generalization of the algorithm to the multiagent paradigm. Like in \cite{luis2021multiagent}, this approach can be applied to multiple cooperative agents using the same uncertainty map to efficiently collect information. In this regard, it is important to address the question of how different agents can effectively cooperate to cover the scenario. This raises the problem of not only static obstacle avoidance, but also between-agent collision avoidance. A future use of multiagent Q-censoring will be necessary.}

\section{Acknowledgments}

This work has been partially funded by the Spanish “Ministerio de Ciencia, Innovación y Universidades, Programa Estatal de I+D+i Orientada a los Retos de la Sociedad” under the Project “Despliegue Adaptativo de Vehículos no Tripulados para Gestión Ambiental en Escenarios Dinámicos RTI2018-098964-B-I00”, and under the PhD grant FPU-2020 (Formación del Profesorado Universitario) of Samuel Yanes Luis, by the regional government Junta de Andalucía under the Projects “Despliegue Inteligente de una red de Vehículos Acuáticos no Tripulados para la monitorización de Recursos Hídricos US-1257508” and "Despliegue y Control de una Red Inteligente de Vehículos Autónomos Acuáticos para la Monitorización de Recursos Hídricos Andaluces PY18-RE0009".

\bibliography{cas-refs.bib}
\bibliographystyle{elsarticle-num} 
\end{document}